\documentclass[journal]{IEEEtran}
\usepackage[colorlinks=true, urlcolor=black]{hyperref}
\usepackage{graphicx}
\ifCLASSOPTIONcompsoc
\usepackage[caption=false,font=normalsize,labelfon
t=sf,textfont=sf]{subfig}
\else
\usepackage[caption=false,font=footnotesize]{subfi
g}
\fi
\usepackage{soul}
\usepackage{array}
\usepackage{amsmath}
\usepackage{multirow}
\usepackage{url}
\usepackage{comment}
\usepackage{mathrsfs}
\usepackage{amsmath}
\usepackage{changes}
\usepackage[flushleft]{threeparttable}
\usepackage[noadjust]{cite}
\DeclareMathOperator*{\argmin}{arg\,min}
\ifCLASSINFOpdf
\else
\fi
\begin{document}
\title{Detecting Comma-Shaped Clouds for Severe Weather Forecasting Using Shape and Motion}

\author{Xinye~Zheng, Jianbo~Ye, Yukun~Chen, Stephen~Wistar,
Jia~Li, \textit{Senior Member, IEEE},\\
Jose~A.~Piedra-Fern\'andez, Michael~A.~Steinberg, and James~Z.~Wang
\thanks{Manuscript received September 21, 2017; revised February 20, 2018 and
May 6, 2018; accepted December 12, 2018. 
This work was supported in part by the National Science Foundation under Grant No. 1027854. The primary computational infrastructures used were supported by the NSF under Grant Nos. ACI-0821527 (CyberStar) and ACI-1053575 (XSEDE). The authors are also grateful to the support by the Amazon AWS Cloud Credits for Research Program and the NVIDIA Corporation's GPU Grant Program.
 \textit{(Corresponding authors:
Xinye Zheng and James Z. Wang)}}
\thanks{X. Zheng, J. Ye, Y. Chen, J. Li and J. Wang are with The Pennsylvania State University, University Park, PA 16802, USA. J. Ye is currently with Amazon Lab126. (e-mails: {\tt \{xvz5220,jwang\}@psu.edu}).}
\thanks{S. Wistar and M. Steinberg are with Accuweather Inc., State College, PA 16803, USA.}
\thanks{J. Piedra-Fern\'andez is with the Department of Informatics, University of Almer\'ia, Almer\'ia 04120, Spain.}
\thanks{Color versions of one or more of the figures in this paper are available online at http://ieeexplore.ieee.org.}
}

\maketitle
\begin{abstract}
Meteorologists use shapes and movements of clouds in satellite images as indicators of several major types of severe storms. Yet, because satellite image data are in increasingly higher resolution, both spatially and temporally, meteorologists cannot fully leverage the data in their forecasts. Automatic satellite image analysis methods that can find storm-related cloud patterns are thus in demand. We propose a machine-learning and pattern-recognition-based approach to detect ``comma-shaped'' clouds in satellite images, which are specific cloud distribution patterns strongly associated with cyclone formulation. In order to detect regions with the targeted movement patterns, we use manually annotated cloud examples represented by both shape and motion-sensitive features to train the computer to analyze satellite images. Sliding windows in different scales ensure the capture of dense clouds, and we implement effective selection rules to shrink the region of interest among these sliding windows. Finally, we evaluate the method on a hold-out annotated comma-shaped cloud dataset and cross-match the results with recorded storm events in the severe weather database. The validated utility and accuracy of our method suggest a high potential for assisting meteorologists in weather forecasting.
\end{abstract}

\begin{IEEEkeywords}
Severe weather forecasting, comma-shaped cloud, Meteorology, satellite images, pattern recognition, AdaBoost.
\end{IEEEkeywords}

\IEEEpeerreviewmaketitle
\section{Introduction}~\label{sec:intro}
Severe weather events such as thunderstorms cause significant losses in property and lives. Many countries and regions suffer from storms regularly, leading to a global issue. For example, severe storms kill over 20 people per year in the U.S.~\cite{avg_storm_kill}. The U.S. government has invested more than 0.5 billion dollars~\cite{noaa_budget_2016} on research to detect and forecast storms, and it has invested billions for modern weather satellite equipment with high-definition cameras.

The fast pace of developing computing power and increasingly higher definition satellite images necessitate a re-examination of conventional efforts regarding storm forecast, such as bare eye interpretation of satellite images~\cite{human_predict_storm}. Bare eye image interpretation by experts requires domain knowledge of cloud involvements and, for a variety of reasons, may result in omissions or delays of extreme weather forecasting. Moreover, the enhancements from the latest satellites which deliver images in real-time at a very high resolution demand tight processing speed. These challenges encourage us to explore how applying modern learning schema on forecasting storms can aid meteorologists in interpreting visual clues of storms from satellite images.

\begin{figure}[htp]\centering
\includegraphics[width=0.4\textwidth,trim=0cm 5cm 3cm 1cm,clip]{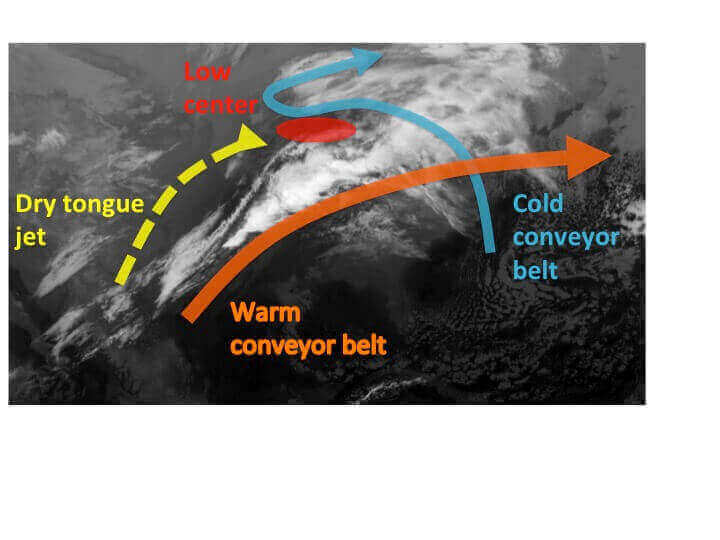}
\caption{An example of the satellite image with the comma-shaped cloud in the north hemisphere. This image is taken at 03:15:19 UTC on Dec 15, 2011 from the fourth channel of the GOES-N weather satellite.}
\label{fig:comma}
\end{figure}

Satellite images with the cyclone formation in the mid-latitude area show clear visual patterns, known as the comma-shaped cloud pattern~\cite{carlson1980airflow}. This typical cloud distribution pattern is strongly associated with mid-latitude cyclonic storm systems. Figure~\ref{fig:comma} shows an example comma-shaped cloud in the Northern Hemisphere, where the cloud shield has the appearance of a comma. Its ``tail'' is formed with the warm conveyor belt extending to the east, and ``head'' within the range of the cold conveyor belt.
The dry-tongue jet forms a cloudless zone between the comma head and the comma tail. 
The comma-shaped clouds also appear in the Southern Hemisphere, but they form towards the opposite direction ({\it i.e.}, an upside-down comma shape). This cloud pattern gets its name because the stream is oriented from the dry upper troposphere and has not achieved saturation before ascending over the low-pressure center. The comma-shaped cloud feature is strongly associated with many types of extratropical cyclones, including hail, thunderstorm, high winds, blizzards, and low-pressure systems. Consequently, we can observe severe events like ice, rain, snow, and thunderstorms around this visual feature~\cite{reed1979cyclogenesis}.  

To capture the comma-shaped cloud pattern accurately, meteorologists have to read different weather data and many satellite images simultaneously, leading to inaccurate or untimely detection of suspected visual signals. Such manual procedures prevent meteorologists from leveraging all available weather data, which increasingly are visual in form and have high resolution. Negligence in the manual interpretation of weather data can lead to serious consequences. Automating this process, through creating intelligent computer-aided tools, can potentially benefit the analysis of historical data and make meteorologists' forecasting efforts less intensive and more timely. This philosophy is persuasive in the computer vision and multimedia community, where images in modern image retrieval and annotation systems are indexed by not only metadata, such as author and timestamp, but also semantic annotations and contextual relatedness based on the pixel content~\cite{li2003automatic,li2008real}.  

We propose a machine-learning and pattern-recognition-based approach to detect comma-shaped clouds from satellite images. The comma-shaped cloud patterns, which have been manually searched and indexed by meteorologists, can be automatically detected by computerized systems using our proposed approach. We leverage the large satellite image dataset in the historical archive to train the model and demonstrate the effectiveness of our method in a manually annotated comma-shaped cloud dataset. Moreover, we demonstrate how this method can help meteorologists to forecast storms using the strong connection of comma-shaped cloud and storm formation.

While all comma-shaped clouds resemble the shape of a comma mark to some extent, the appearance and size of one such cloud can be very different from those of another. This makes conventional object detection or pattern matching techniques developed in computer vision inappropriate because they often assume a well-defined object shape ({\it e.g.} a face) or pattern ({\it e.g.} the skin texture of a zebra).

The key visual cues that human experts use in distinguishing comma-shaped clouds are {\it shape} and {\it motion}. During the formulation of a cyclone, the ``head'' of the comma-shaped cloud, which is the northwest part of the cloud shield, has a strong rotation feature. The dense cloud patch forms the shape of a comma, which distinguishes the cloud patch from other clouds. To emulate meteorologists, we propose two novel features that consider both shape and motion of the cloud patches, namely, \textit{Segmented HOG} and \textit{Motion Correlation} Histogram, respectively. We detail our proposals in Sec.~\ref{sec:segmentation} and Sec.~\ref{sec:corr}. 

Our work makes two main contributions. First, we propose novel shape and motion features of the cloud using computer vision techniques. These features enable computers to recognize the comma-shaped cloud from satellite images. Second, we develop an automatic scheme to detect the comma-shaped cloud on the satellite images. Because the comma-shaped cloud is a visual indicator of severe weather events, our scheme can help meteorologists forecast such events.

\subsection{Related Work}

Cloud segmentation is an important method for detecting storm cells. Lakshmanan {\it et al.}~\cite{lakshmanan2003multiscale} proposed a hierarchical cloud-texture segmentation method for satellite image. Later, they improved the method by applying watershed transform to the segmentation and using pixel intensity thresholding to identify storms~\cite{lakshmanan2009efficient}. However, brightness temperature in a single satellite image is easily affected by lighting conditions, geographical location, and satellite imager quality, which is not fully considered in the thresholding-based methods. Therefore, we consider these spatial and temporal factors and segment the high cloud part based on the Gaussian Mixture Model (GMM).

Cloud motion estimation is also an important method for storm detection, and a common approach estimates cloud movements through cross-correlation over adjacent images.
Some earlier work~\cite{leese1970determination} and \cite{smith1972automated} applied the cross-correlation method to derive the motion vectors from cloud textures, which was later extended to multi-channel satellite images in~\cite{evans2006cloud}. The cross-correlation method could partly characterize the airflow dynamics of the atmosphere and provide meaningful speed and direction information on large areas~\cite{johnson1998storm}. After being introduced in the radar reflectivity images, the method was applied in the automatic cloud-tracking systems using satellite images. A later work~\cite{carvalho2001satellite} implemented the cross-correlation in predicting and tracking the Mesoscale Convective Systems (MCS, a type of storms). Their motion vectors were computed by aggregating nearby pixels at two consecutive frames; thus, they are subject to spatially smoothed effects and miss fine-grained details. Inspired by the ideas of motion interpretation, we define a novel correlation aiming to recognize cloud motion patterns in a longer period. The combination of motion and shape features demonstrates high classification accuracy on our manually labeled dataset.

Researchers have applied pattern recognition techniques to interpret storm formulation and movement extensively. Before the satellite data reached a high resolution, earlier works constructed storm formulation models based on 2D radar reflectivity images in the 1970s. The primary techniques can be categorized into \textit{cross correlation}~\cite{rinehart1978three} and \textit{centroid tracking}~\cite{crane1979automatic} methods. According to the analysis, cross-correlation based methods are more capable of accurate storm speed estimation, while centroid-tracking-based methods are better at tracking isolated storm cells. 

Taking advantages of these two ideas, Dixon and Wiener developed the renowned centroid-based storm nowcasting algorithm, named Thunderstorm Identification, Tracking, Analysis and Nowcasting (TITAN)~\cite{dixon1993titan}. This method consists of two steps: identifying the isolated storm cells and forecasting possible centroid locations. Compared with former methods, TITAN can model and track some storm merging and splitting cases. This method, however, can have large errors if the cloud shape changes quickly~\cite{han20093d}. Some later works attempted to model the storm identification process mathematically. For instance, ~\cite{lakshmanan2009data} and~\cite{lopez2009discriminant} used statistical features of the radar reflection to classify regions into storm or storm-less classes. 

Recently, Kamani {\it et al.} proposed a severe storm detection method by matching the skeleton feature of bow echoes ({\it i.e.}, visual radar patterns associated with storms) on radar images in~\cite{kamani2016shape}, with an improvement presented in~\cite{kamani2017skeleton}. The idea is inspiring, but radar reflectivity images have some weaknesses in extreme weather precipitation~\cite{westrick1999limitations}. First, the distribution of radar stations in the contiguous United States (CONUS) is uneven. The quality of ground-based radar reflectivity data is affected by the distance to the closest radar station to some extent. Second, detections of marine events are limited because there are no ground stations in the oceans to collect reflectivity signals. Finally, severe weather conditions would affect the accuracy of radar. Since our focus is on severe weather event detection, radar information may not provide enough timeliness and accuracy for detection purposes.

Compared with the weather radar, multi-channel geosynchronous satellites have larger spatial coverages and thus are capable of providing more global information to the meteorologists. Take the infrared spectral channel in the satellite imager as an example: the brightness of a pixel reflects the temperature and the height of the cloud top position~\cite{weinreb2011conversion}, which in turn provides the physical condition of the cloud patch at a given time. To find more information about the storm, researchers have applied many pattern recognition methods to satellite data interpretation, like combining multiple channels of image information from the weather satellite~\cite{evans2006cloud} and combining images from multiple satellites~\cite{ho2008automated}. Image analysis methods, including cloud patch segmentation and background extraction~\cite{lakshmanan2003multiscale}~\cite{srivastava2003onboard}, cyclone identification~\cite{lee2000tropical}~\cite{ho2008automated2}, cloud motion estimation~\cite{zhou2001tracking}, and vortex extraction~\cite{zhang2014locating}~\cite{zhang2017severe}, have also been incorporated in severe weather forecasting from satellite data. However, these approaches lack an attention mechanism that can focus on areas most likely to have major destructive weather conditions. Most of these methods do not consider high-level visual patterns ({\it i.e.} larger patterns spatially) to describe the severe weather condition. Instead, they represent extreme weather phenomena by relatively low-level image features.

\subsection{Proposed Spatiotemporal Modeling Approach}

In contrast to current technological approaches, meteorologists, who have geographical knowledge and rich experience of analyzing past weather events, typically take a top-down approach. They make sense of available weather data in a more global (in contrast to local) fashion than numerical simulation models do.
For instance, meteorologists can often make reasonable judgments about near-future weather conditions by looking at the general cloud patterns and the developing trends from satellite image sequences, while existing pattern recognition methods in weather forecasting do not capture such high-level clues such as comma-shaped clouds. Unlike the conventional object-detection task, detecting comma-shaped clouds is highly challenging. First, some parts of cloud patches can be missing from satellite images. Second, such clouds vary substantially in terms of scales, appearance, and moving trajectory. Standard object detection techniques and their evaluation methods are inappropriate. 

To address these issues, we propose a new method to detect the comma-shaped cloud in satellite images. Our framework implements computer vision techniques to design the task-dependent features, and it includes re-designed data-processing pipelines. Our proposed algorithm can effectively identify comma-shaped clouds from satellite images. In the Evaluation and Case Study sections, we show that our method contributes to storm forecasting using real-world data, and it can produce earlier and more sensitive detections than human perception in some cases. 

The remainder of the paper is organized into four sections. Section~\ref{sec:dataset} describes the satellite image dataset and the training labels. Sec.~\ref{sec:method} details our machine learning framework, with the evaluation results in Sec.~\ref{sec:result}. We provide some case studies in Sec.~\ref{sec:casestudy} and draw conclusions in Sec.~\ref{sec:conclude}.

\begin{figure}[tbp]\centering
\includegraphics[width=0.47\textwidth,trim=21.5cm 2.3cm 8cm 2.5cm,clip]{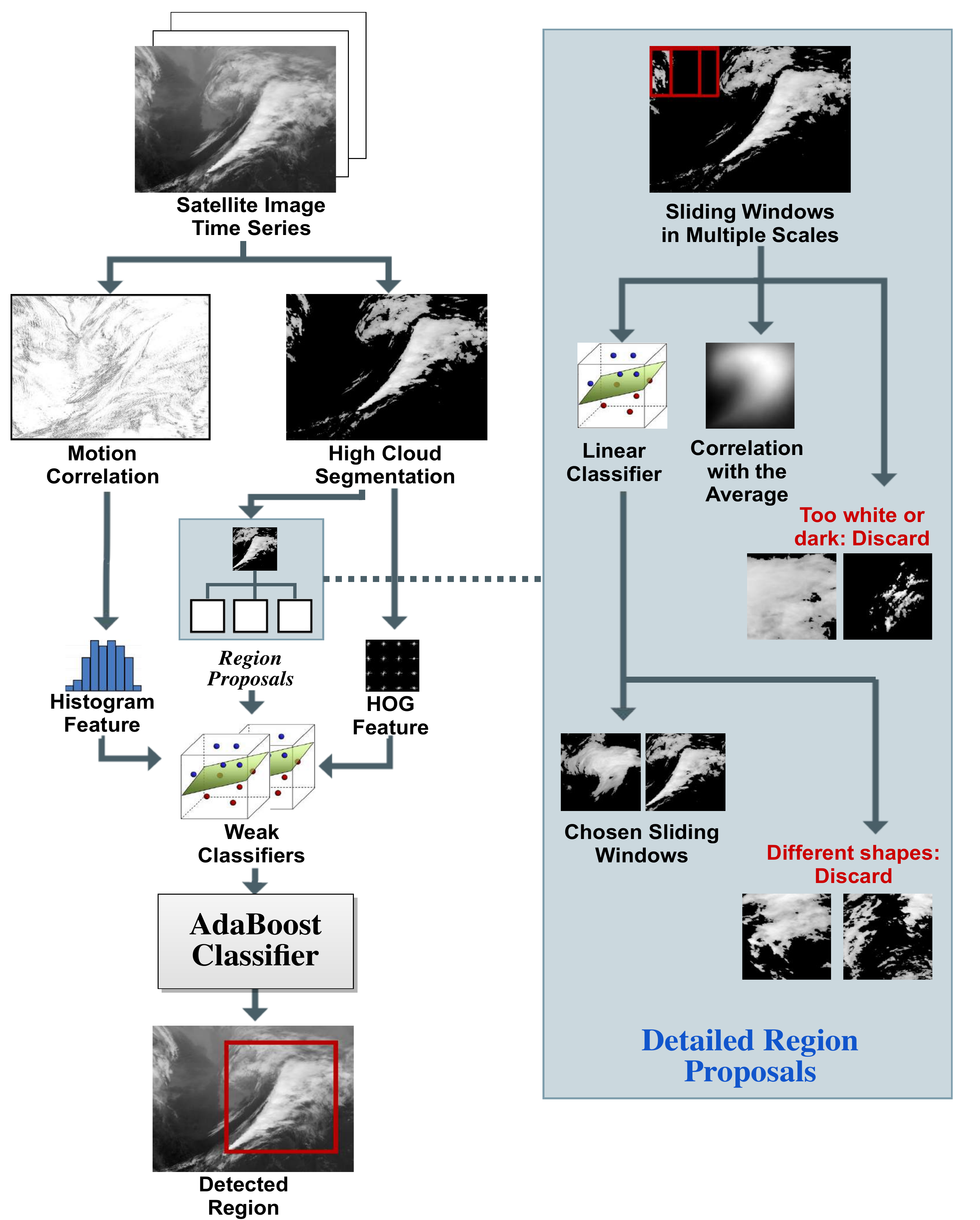}
\caption{\textit{Left:} The pipeline of the comma-shaped cloud detection process. high-cloud segmentation, region proposals, correlation with motion prior, constructions of weak classifiers, and the AdaBoost detector are described in Sections~\ref{sec:segmentation},~\ref{sec:corr},~\ref{sec:regions},~\ref{sec:weakclassifier}, and~\ref{sec:adaboost}, respectively. \textit{Right:} The detailed selection process for region proposals.}
\label{fig:framework}
\end{figure}

\section{The Dataset}~\label{sec:dataset}

Our dataset consists of the GOES-M weather satellite images for the year 2008 and the GOES-N weather satellite images for the years 2011 and 2012. We select these three years because the U.S. experienced more severe thunderstorm activities than it does in a typical year. GOES-M and GOES-N weather satellites are in the geosynchronous orbit of Earth and provide continuous monitoring for intensive data analysis. Among the five channels of the satellite imager, we adopt the fourth channel, because it is infrared among the wavelength range of (10.20 - 11.20$\mu$m), and thus can capture objects of meteorological interest including clouds and sea surface~\cite{arking1985retrieval}. The channel is at the resolution of 2.5 miles and the satellite takes pictures of the northern hemisphere at the 15th minute and the 45th minute of each hour.
We use these satellite frames of CONUS at 20$^{\circ}$-50$^{\circ}$ N, 60$^{\circ}$-120$^{\circ}$ W. Each satellite image has 1,024$\times$2,048 pixels, and a gray-scale intensity that positively correlates with the infrared temperature. After the raw data are converted into the image data in accordance with the information in~\cite{earth1998location}, each image pixel represents a specific geospatial location.

The labeled data of this dataset consist of two parts, (1) comma-shaped clouds identified with the help of meteorologists from AccuWeather Inc., and (2) an archive of storm events for these three years in the U.S.~\cite{stormeventdatabase}.

\begin{figure*}[!tbp]\centering
\begin{minipage}{0.47\linewidth}
\includegraphics[width=\textwidth,trim=4.5cm 0.2cm 1cm 4cm,clip]{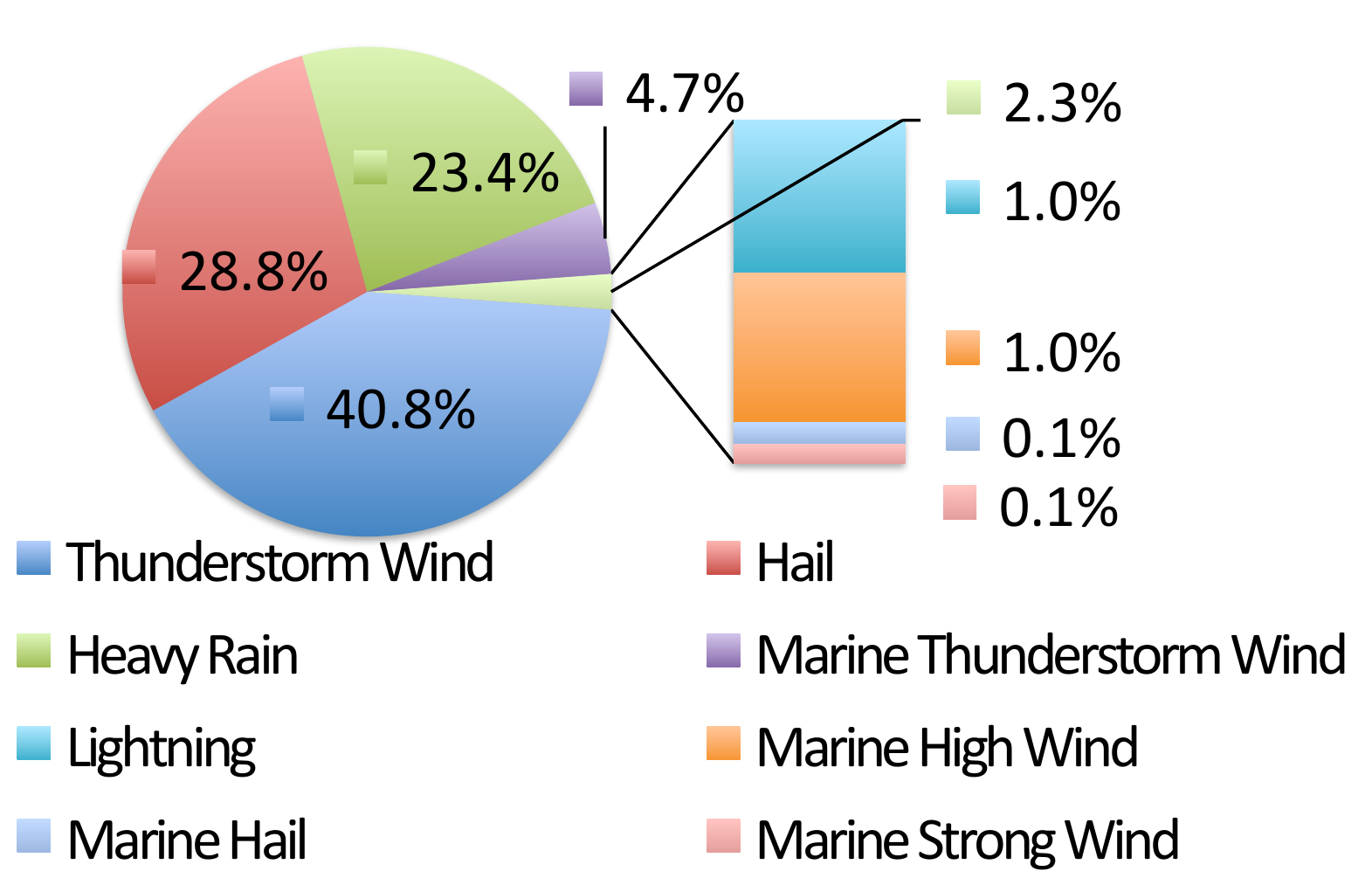}
\end{minipage}
\begin{minipage}{0.47\linewidth}
\includegraphics[width=0.49\textwidth,trim=5cm 4.1cm 4.5cm 5cm,clip]{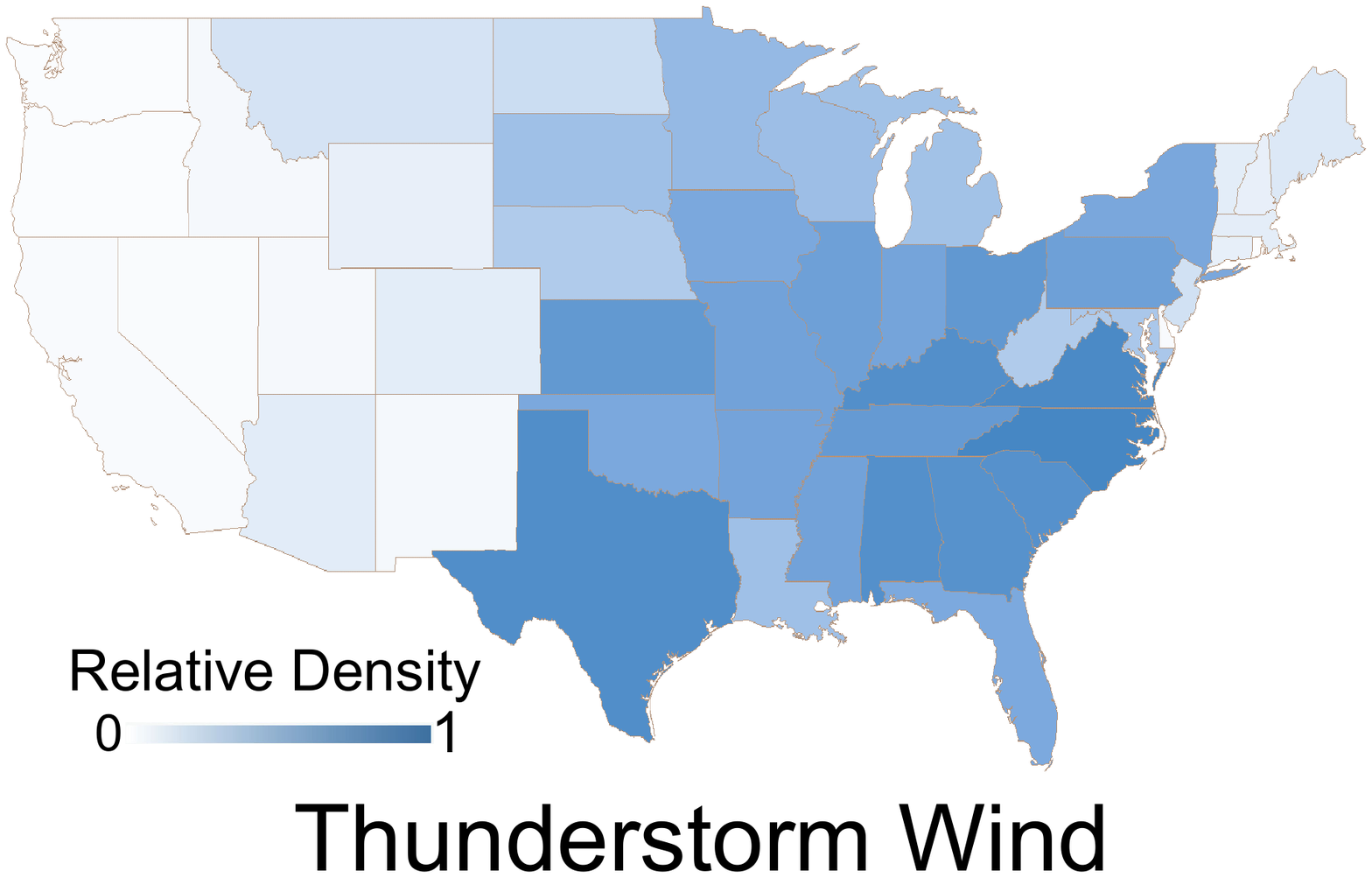}
\includegraphics[width=0.49\textwidth,trim=5cm 4.1cm 4.5cm 5cm,clip]{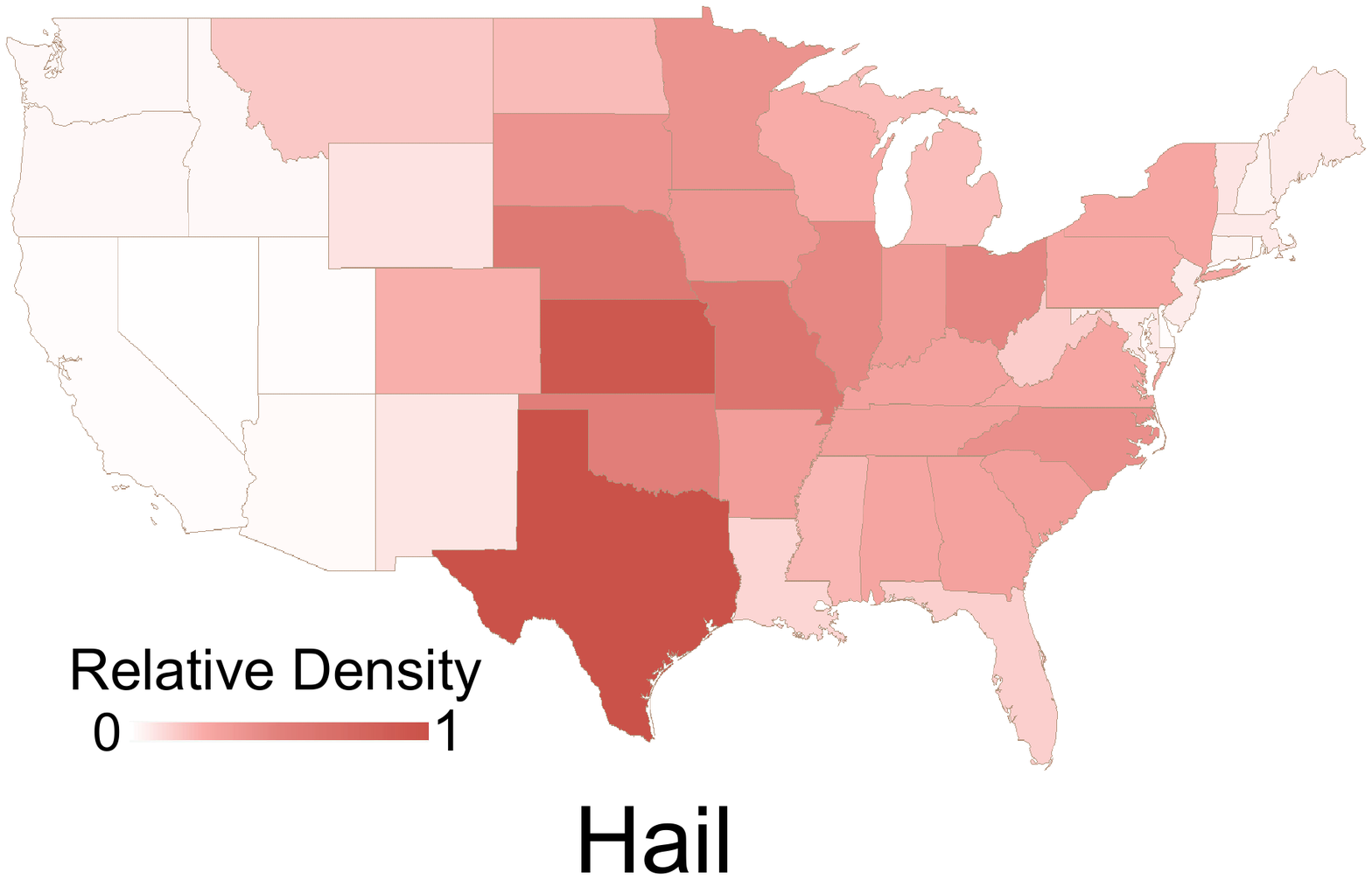}\\
\includegraphics[width=0.49\textwidth,trim=5cm 4.1cm 4.5cm 5cm,clip]{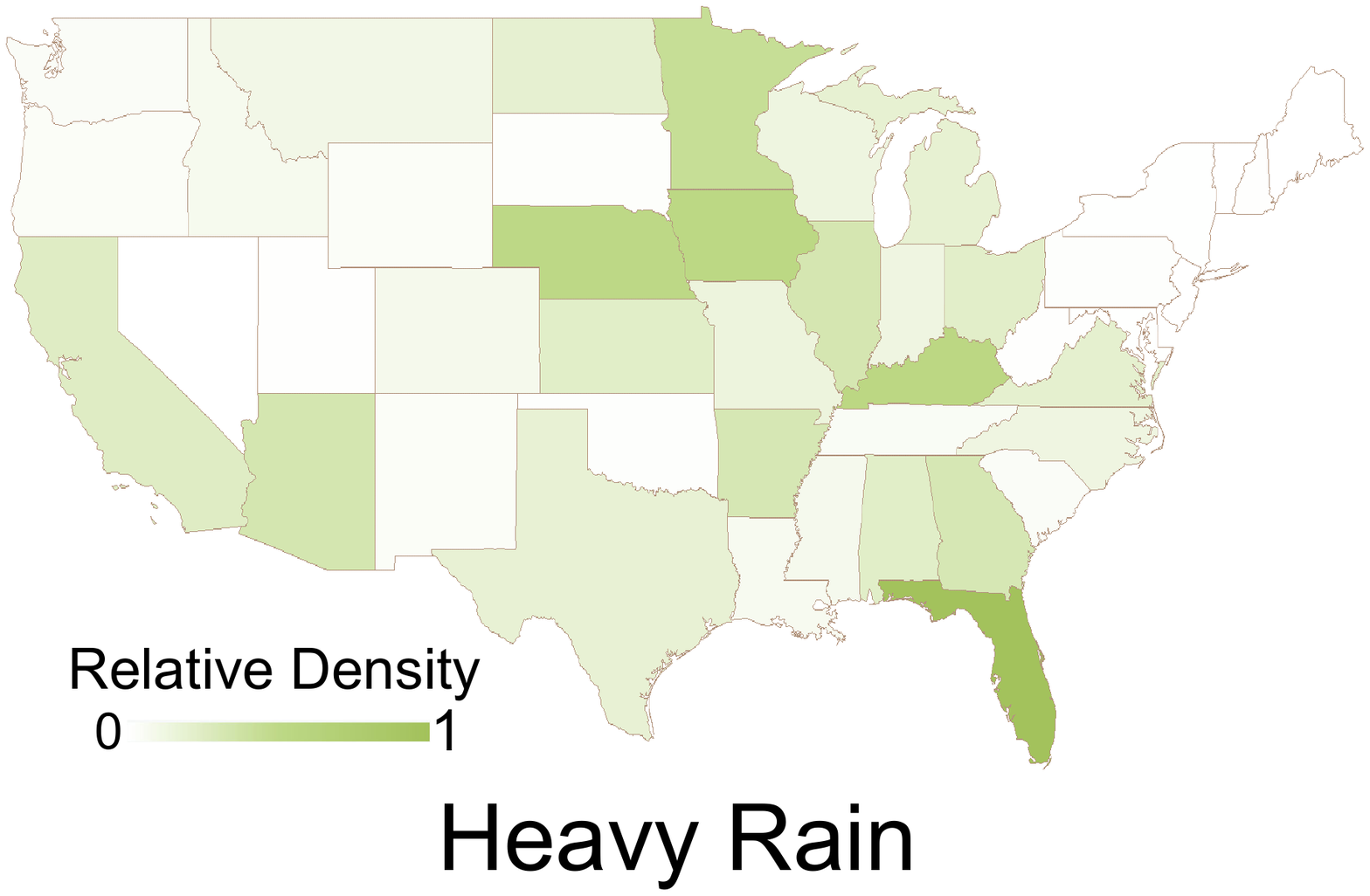}
\includegraphics[width=0.49\textwidth,trim=5cm 4.1cm 4.5cm 5cm,clip]{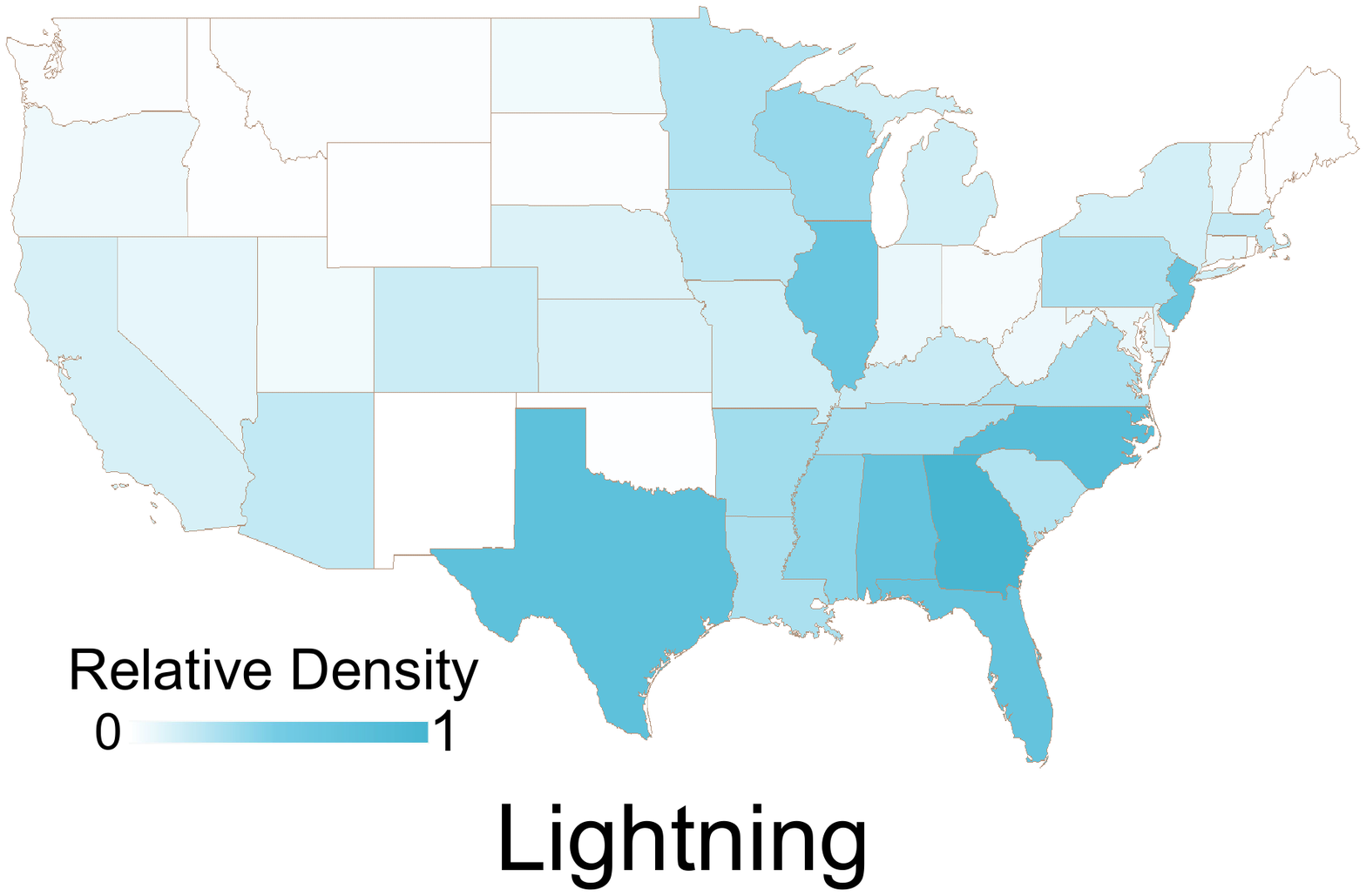}
\end{minipage}
\caption{Proportions and geographical distributions of different severe weather events in the year 2011 and 2012. \textit{Left:} Proportions of different categories of selected storm types. \textit{Right:} State-wise geographical distributions of land-based storms.}
\label{fig:storm_record}
\end{figure*}

To create the first part of the annotated data, we manually label comma-shaped clouds by using tight squared bounding boxes around each such cloud. If a comma-shaped cloud moves out of the range, we ensure that the head and tail of the comma are in the middle part of the square. The labeled comma-shaped clouds have different visual appearances, and their coverage varies from a width of 70 miles to 1,000 miles. Automatic detection of them is thus nontrivial. The labeled dataset includes a total of 10,892 comma-shaped cloud frames in 9,321 images for the three years 2008, 2011, and 2012. Most of them follow the earlier description of comma-shaped clouds, with the visible rotating head part, main body heading from southwest to northeast, and the dark dry slot area between them.

The second part of the labeled data consists of storm observations with detailed information, including time, location, range, and type. Each storm is represented by its latitude and longitude in the record. We ignore the range differences between storms because the range is relatively small ($<$ 5 miles) compared with our bounding boxes (70 $\sim$ 1000 miles). Every event recorded in the database had enough severity to cause death, injuries, damage, and disruption to commerce. The total estimated damage from storm events for the years 2011 and 2012 surpassed two billion dollars~\cite{annual2011summary}. From the database, we chose eight types of severe weather records\footnote{Tornadoes are included in the Thunderstorm Wind type.} that are known to correlate strongly with the comma-shaped clouds and happen most frequently among all types of events. The distribution of these eight types of severe weather events is shown in the left part of Fig.~\ref{fig:storm_record}. Among those eight types of severe weather events, thunderstorm winds, hail, and heavy rain happen most frequently ($\sim 93 \%$ of the total events). The state-wise geographical distributions of some types of storm events are in the right half of Fig.~\ref{fig:storm_record}. Because marine-based events do not have associated state information, we only visualize the geographical distributions for land-based storm events. With the exception of heavy rains, these severe weather events happen more frequently in East CONUS.

In our experiments, we include only storms that lasted for more than 30 minutes because they overlapped with at least one satellite image in the dataset. Consequently, we have 5,412 severe storm records for the years 2011 and 2012 in the CONUS area for testing purpose, and their time span varies from 30 minutes to 28 days. 

\section{Our Proposed Detection Method}~\label{sec:method}

Fig.~\ref{fig:framework} shows our proposed comma-based cloud detection pipeline framework. We first segment the cloud from the background in Sec.~\ref{sec:segmentation}, and then we extract shape and motion features of clouds in Sec.~\ref{sec:corr}. Well-designed region proposals in Sec.~\ref{sec:regions} shrink the searching range on satellite images. The features on our extracted region proposals are fed into weak classifiers in Sec.~\ref{sec:weakclassifier} and then we ensemble these weak classifiers as our comma-shaped cloud detector in Sec.~\ref{sec:adaboost}. We now detail the technical setups in this section. 

\subsection{High-Cloud Segmentation}\label{sec:segmentation}

We first segment the high cloud part from the noisy original satellite data.
Raw satellite images contain all the objects that can be seen from the geosynchronous orbit, including land, seas, and clouds. Among all the visible objects in satellite images, we focus on the dense middle and top clouds, which we refer to as ``high cloud'' in the following. The high cloud is important because the comma-shaped phase is most evident in this part, according to~\cite{carlson1980airflow}. 

The prior work~\cite{otsu1979threshold} implemented the single-threshold segmentation method to separate clouds from the background. This method is based on the fact that high cloud looks brighter than other parts of the infrared satellite images~\cite{weinreb2011conversion}.
We evaluate this method and show the result in the second column of Fig.~\ref{fig:compare}.
Although this method can segment most high clouds from the background, it misses some cloud boundaries. Because Earth has periodic temperature changes and ground-level temperature variations, and the variations are affected by many factors including terrains, elevation, and latitudes, a single threshold cannot adapt to all these cases.

The imperfection of the prior segmentation method motivates us to explore a data-driven approach. The overall idea of the new segmentation scheme is described as follows: To be aware of spatiotemporal changes of the satellite images, we divide the image pixels into tiles, and then model the samples of each unit using a GMM. Afterward, we identify the existence of a component that most likely corresponds to the variations of high cloud-sky brightness.

We build independent GMMs for each hour and each spatial region to address the challenges of periodic sunlight changes and terrain effects. As sunlight changes in a cycle of one day, we group satellite images by hours and estimate GMMs for each hour separately. Furthermore, since land conditions also affect light conditions, we divide each satellite image into non-overlapping windows according to their different geolocations. Suppose all the pixels are indexed by their time stamp $t$ and spatial location $\mathbf x$, we divide each day into 24 hours, and divide each image into non-overlapping windows. Each window is a square of $32\times 32$. Thus, for each hour $h$ and each window $L$, {\it i.e.}, $T_h\times X_L$, we form a group of pixels $G_{h,L}=\{I(t,\mathbf x): t\in T_h, \mathbf x\in X_L\} = \{I_{h,L}(t,\mathbf x)\}$, with brightness $I(t,\mathbf x)\in$ [0, 255]. Each pixel group $G_{h,L}$ has about 150,000 samples. We model each group by a GMM with the number of components of that group $K_{h,L}$ to be 2 or 3, {\it i.e.}, 
\begin{equation*}
I_{h,L}(t,\mathbf x) \sim \sum_{i=1}^{K_{h,L}} \varphi^{(i)}_{h,L} \mathcal{N} \left( \mu^{(i)}_{h,L} ,\Sigma^{(i)}_{h,L} \right),
\end{equation*}
where 
\begin{align*}
K_{h,L} & = \argmin_{i = 2,3; (t,\mathbf x) \in T_h \times X_L} \left\{AIC(K_{h,L}=i|t, \mathbf x)\right\}.
\end{align*}
Here AIC($\cdot$) is the Akaike information criterion function of $K_{h,L}$. $\psi_{h,L} = \left\{\varphi^{(i)}_{h,L},\mu^{(i)}_{h,L} ,\Sigma^{(i)}_{h,L}\right\}_{i = 1}^{K_{h,L}}$ are GMM parameters satisfying $\mu^{(i)}_{h,L} > \mu^{(j)}_{h,L}$ for $\forall i > j$, which are estimated by the k-means++ method~\cite{arthur2007k}. We can interpret the GMM component number K = 2 as the GMM peaks fit high-sky clouds and land, while K = 3 as the GMM peaks fit high-sky clouds, low-sky clouds, and the land. So for each GMM $\psi_{h,L}$, the component with the largest mean is the one modeling high cloud-sky. We then compute the normalized density of the point $(t, \mathbf x)$ over $\psi_{h,L}$. We annotate this normalized density as $\left\{p_{h,L}^{(i)}(t,\mathbf x)\right\}_{i = 1}^{K_{h,L}}$ and define the intensity value after segmentation to be 
$\tilde{I}(t,\mathbf x):=$
\begin{equation}
\left\{\begin{array}{ll}
I_{h,L}(t,\mathbf x)\cdot p_{h,L}^{(1)}(t,\mathbf x) & \mbox{if } I_{h,L}(t,\mathbf x)\cdot p_{h,L}^{(1)}(t,\mathbf x) \ge \sigma \\
0 & \mbox{otherwise},
\end{array}\right.
\label{eq:cut}
\end{equation}
where $\sigma$ is chosen empirically between 100 and 130, with low impact to the features extracted. In our experiment, we choose 120 for convenience.

We then apply a min-max filter between neighboring GMMs in spatiotemporal space.
Based on the assumption that cloud movement is smooth in spatiotemporal space, GMM parameters $\psi_{h,L}$ should be a continuous function over $h$ and $L$. For most pixel groups which we have examined, we observe that our segmented cloud changes smoothly. But in case the GMM component number changes, $\mu^{(1)}_{h,L}$ would also change in both $h$ and $L$, resulting in significant changes to the segmented cloud.
To smooth the cloud boundary part, we post-process a min-max filter to update $\mu^{(1)}_{h,L}$, which is given by
\begin{equation}
\mu_{h,L}^{(1),\mbox{new}}:=\max\left\{\mu^{(1)}_{h,L}, {\min}_{\substack{h'\in \mathcal N_h \\L'\in \mathcal M_L}}\{\mu^{(1)}_{h',L'}\}\right\},
\label{eq:minimax_filter}
\end{equation}
where $\mathcal N_h = \left[h-1, h+1 \right]$ and $\mathcal M_L= \left\{l:\left|\ l - L\right|\ \leq 1\right\}$.
The min-max filter leverages smoothness of GMMs within spatiotemporal neighbors. After applying Eq.~\eqref{eq:minimax_filter}, we update normalized densities and receive more smooth results with Eq.~\eqref{eq:cut}. Example high-cloud segmentation results are shown in the third column of Fig.~\ref{fig:compare}. At the end of this step, high clouds are separated with detailed local information, while the shallow clouds and the land are removed.

\subsection{Correlation with Motion Prior}~\label{sec:corr}

Another evident feature of the comma-shaped clouds is motion. In the cyclonic system, the jet stream has a strong trend to rotate around the low center, which makes up the head part of the comma in the satellite image~\cite{crane1979automatic}. We design a visual feature to extract this cloud motion information, namely \textit{Motion Correlation} in this section. The key idea is that the \textit{same} cloud at two \textit{different} spatiotemporal points should have a strong positive correlation in appearance,
based on a reasonable assumption that clouds move at a nearly-uniform speed within a small spatial and time span.
Thus, cloud movement direction can be inferred from the direction of the maximum correlation. This assumption was first applied to compute cross-correlation in~\cite{leese1970determination}.

We therefore define the motion correlation of location $\mathbf x$ on the time interval of $(t-T, t]$ to be: 
\begin{equation}
M(t, \mathbf x) = 
\mbox{corr}_{t_0 \in (t-T, t]} (I(t - t_0,\mathbf x), I(t, \mathbf x + \mathbf h))\;,
\label{eq:motion_corr}
\end{equation}
where $\mbox{corr}(\cdot,\cdot)$ denotes the Pearson correlation coefficient, and $\mathbf h$ is the cloud displacement distance in time interval $T$.
This motion correlation can be viewed as an improved cross-correlation in~\cite{leese1970determination}, which we mentioned in Sec.~\ref{sec:intro}. The cross-correlation can be written in the following form:
\begin{equation}
M_0(t, \mathbf x) = 
\mbox{corr}_{\left \| \mathbf x_1 - \mathbf x \right \|\leq \mathbf h_0} (I(t - T_0,\mathbf x), I(t, \mathbf x_1 + \mathbf h))\;,
\label{eq:motion_ref}
\end{equation}
where $T_0$ is the time span between two successive satellite images.

\begin{figure}[!htbp]
\vspace{-0.16in}
\centering
\subfloat {\includegraphics[width=0.09\textwidth,trim=950 550 650 26,clip]{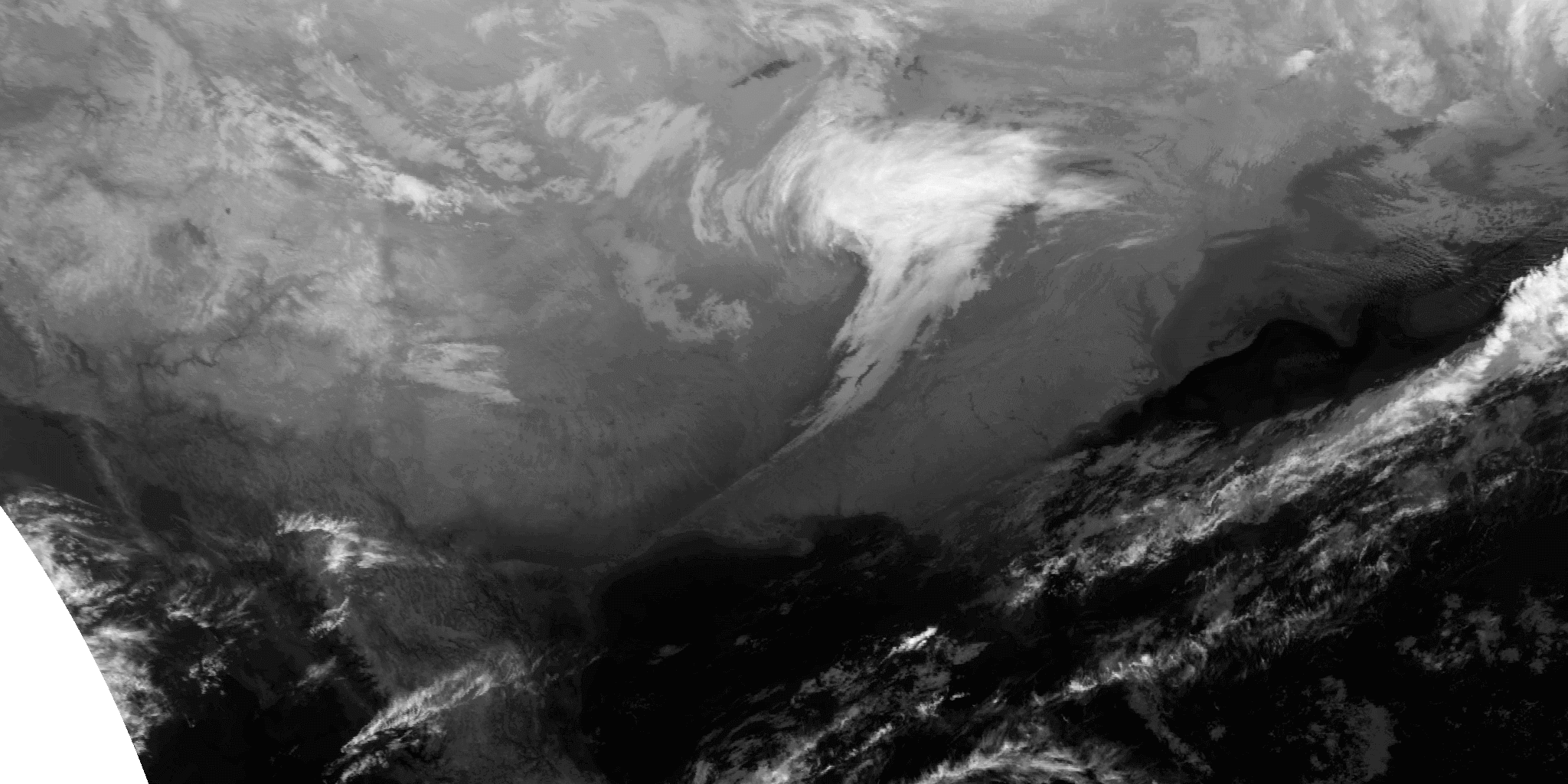}}\hspace{0.01in}
\subfloat {\includegraphics[width=0.09\textwidth,trim=950 550 650 26,clip]{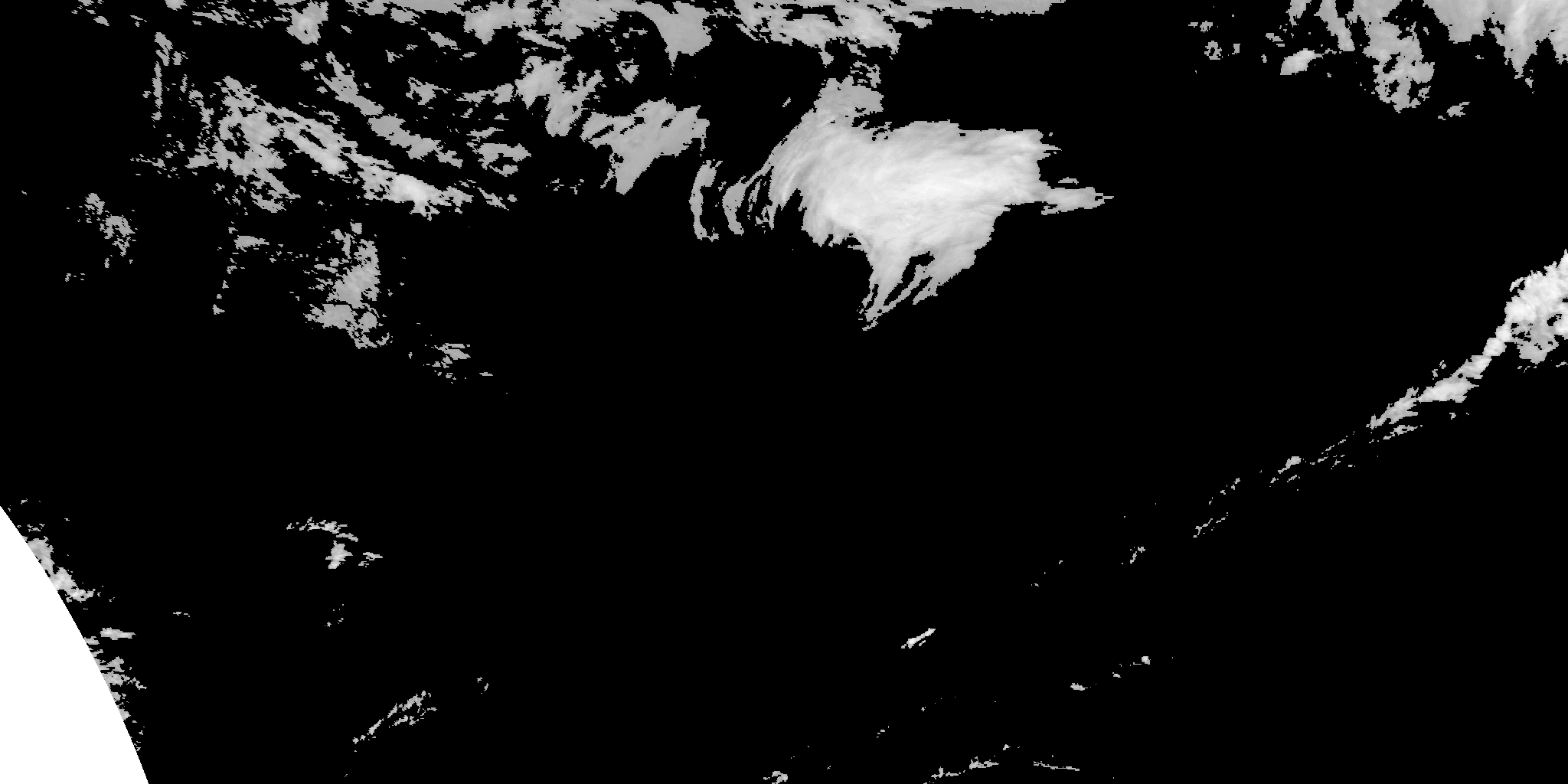}}\hspace{0.01in}
\subfloat {\includegraphics[width=0.09\textwidth,trim=950 550 650 26,clip]{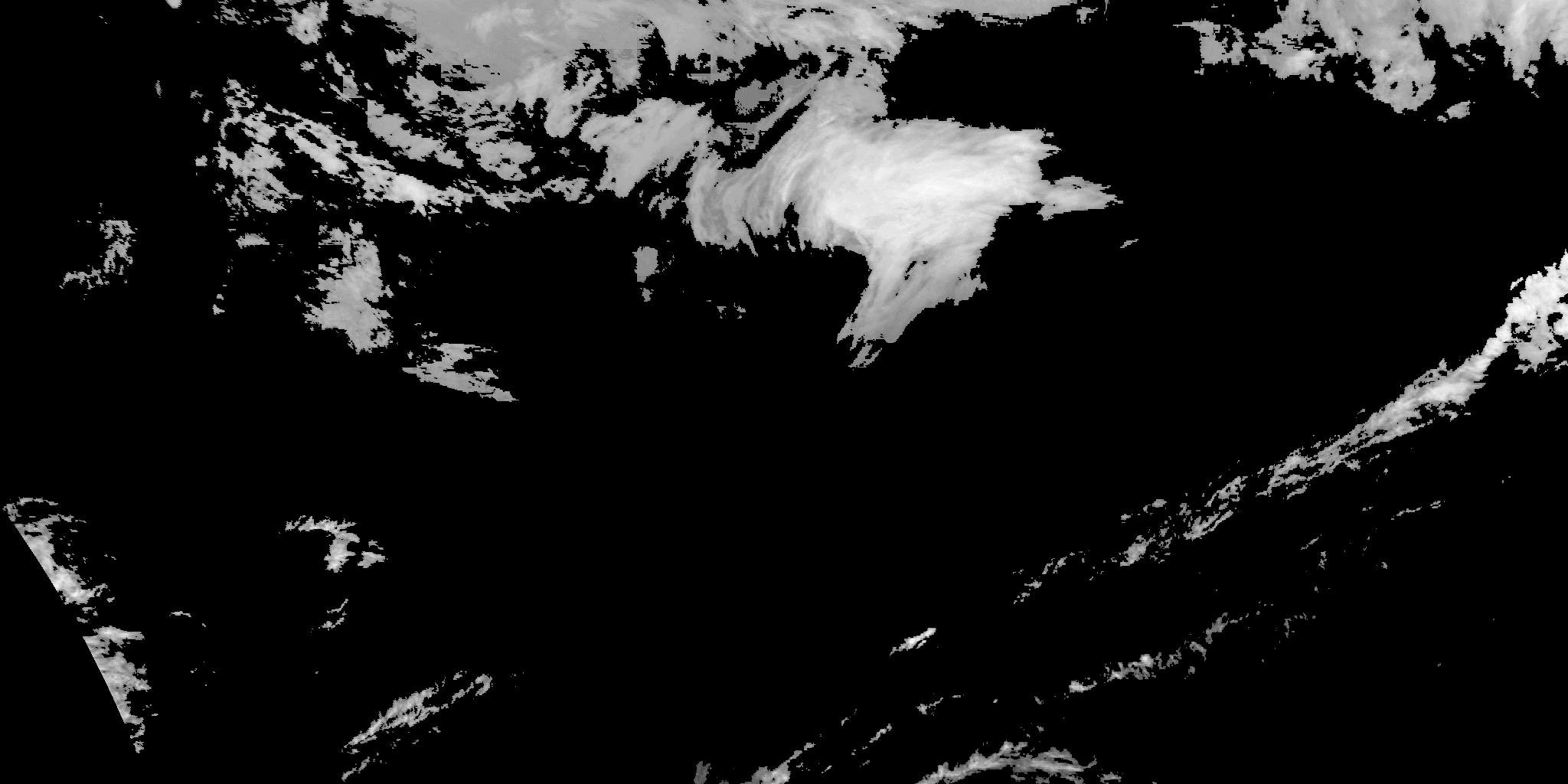}}\hspace{0.01in}
\subfloat {\includegraphics[width=0.09\textwidth,trim=950 550 650 26,clip]{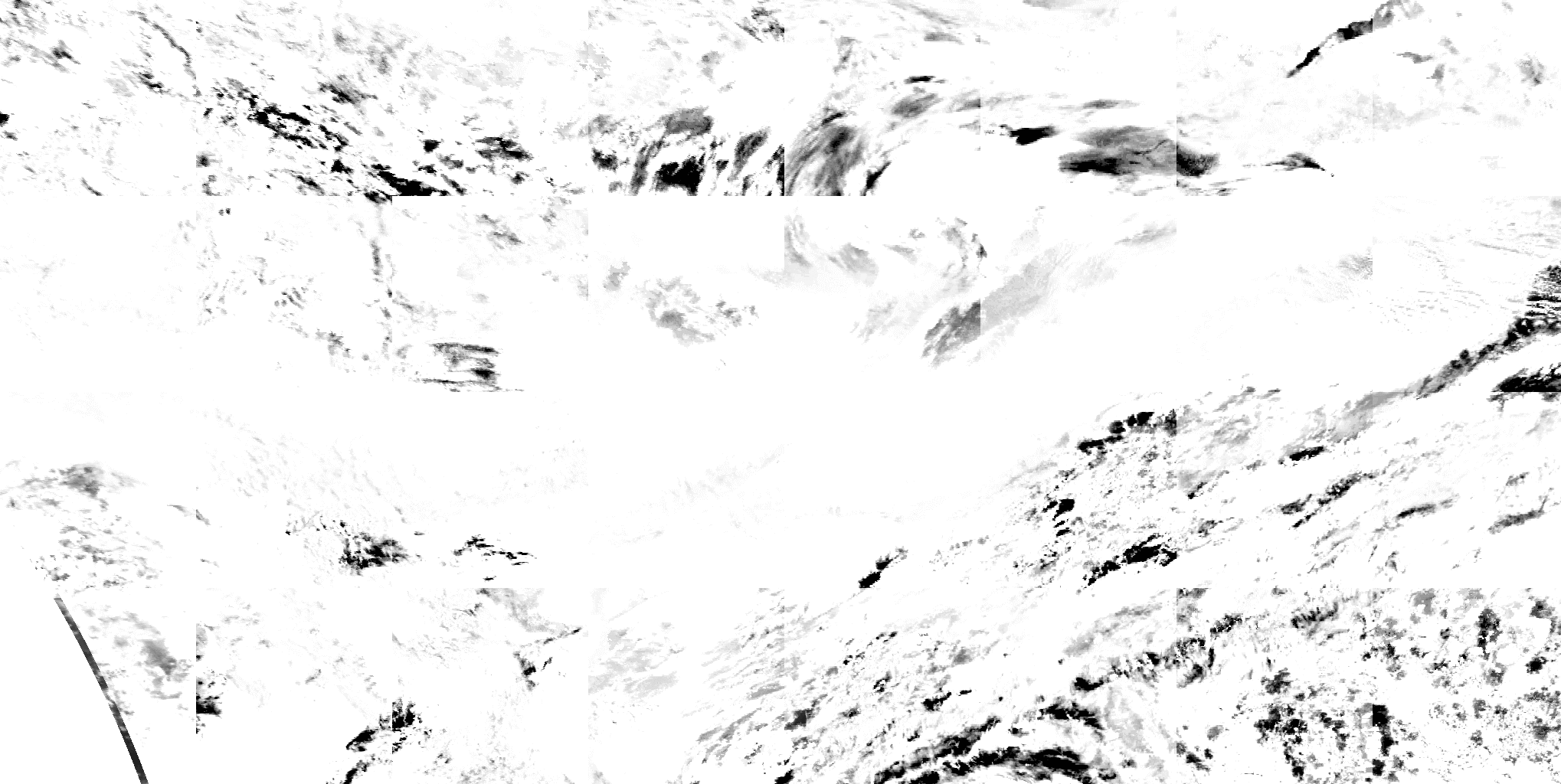}}\hspace{0.01in}
\subfloat {\includegraphics[width=0.09\textwidth,trim=950 550 650 26,clip]{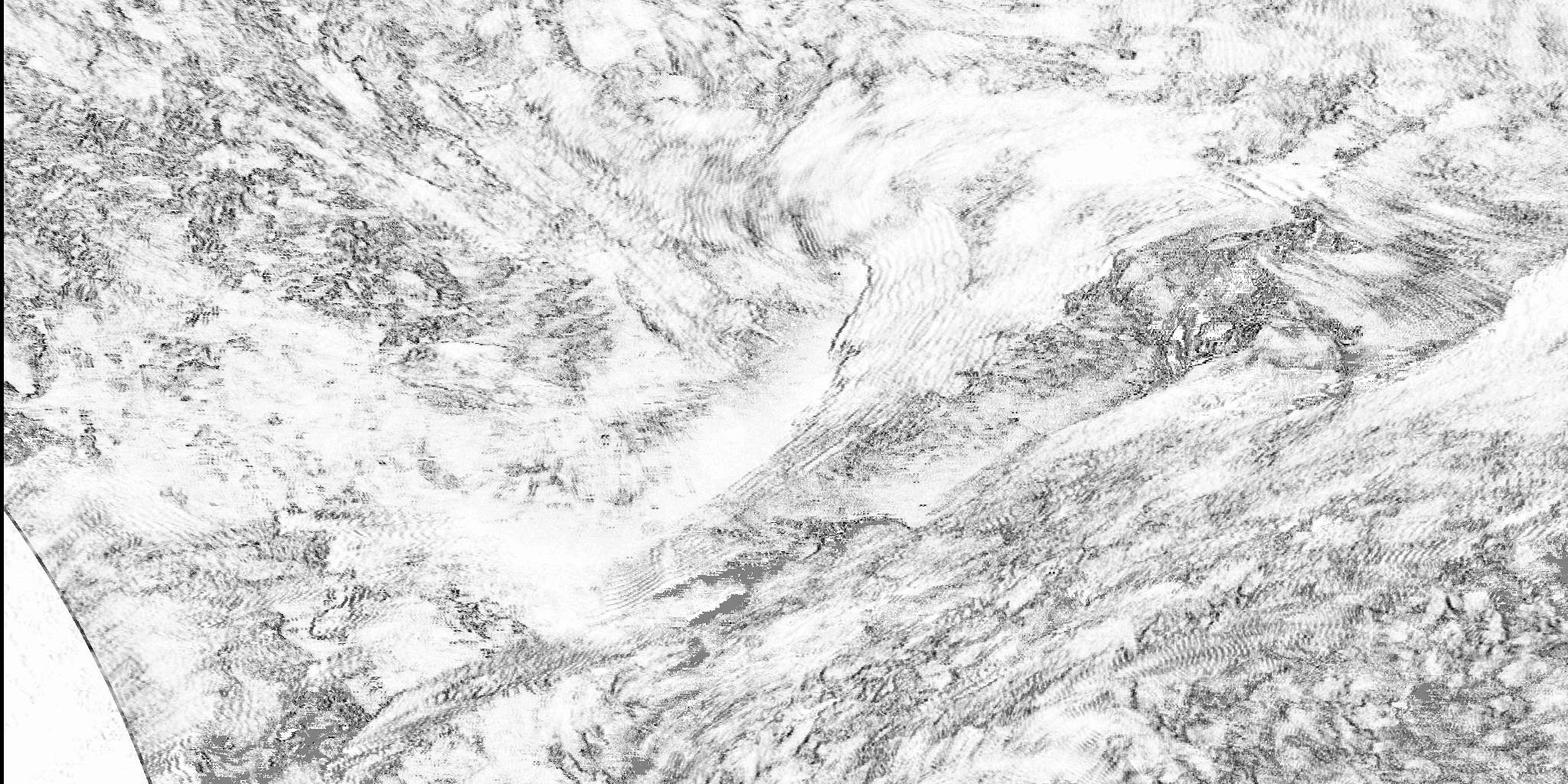}}

\vspace{-0.1in}

\subfloat {\includegraphics[width=0.09\textwidth,trim=150 160 1036 0,clip]{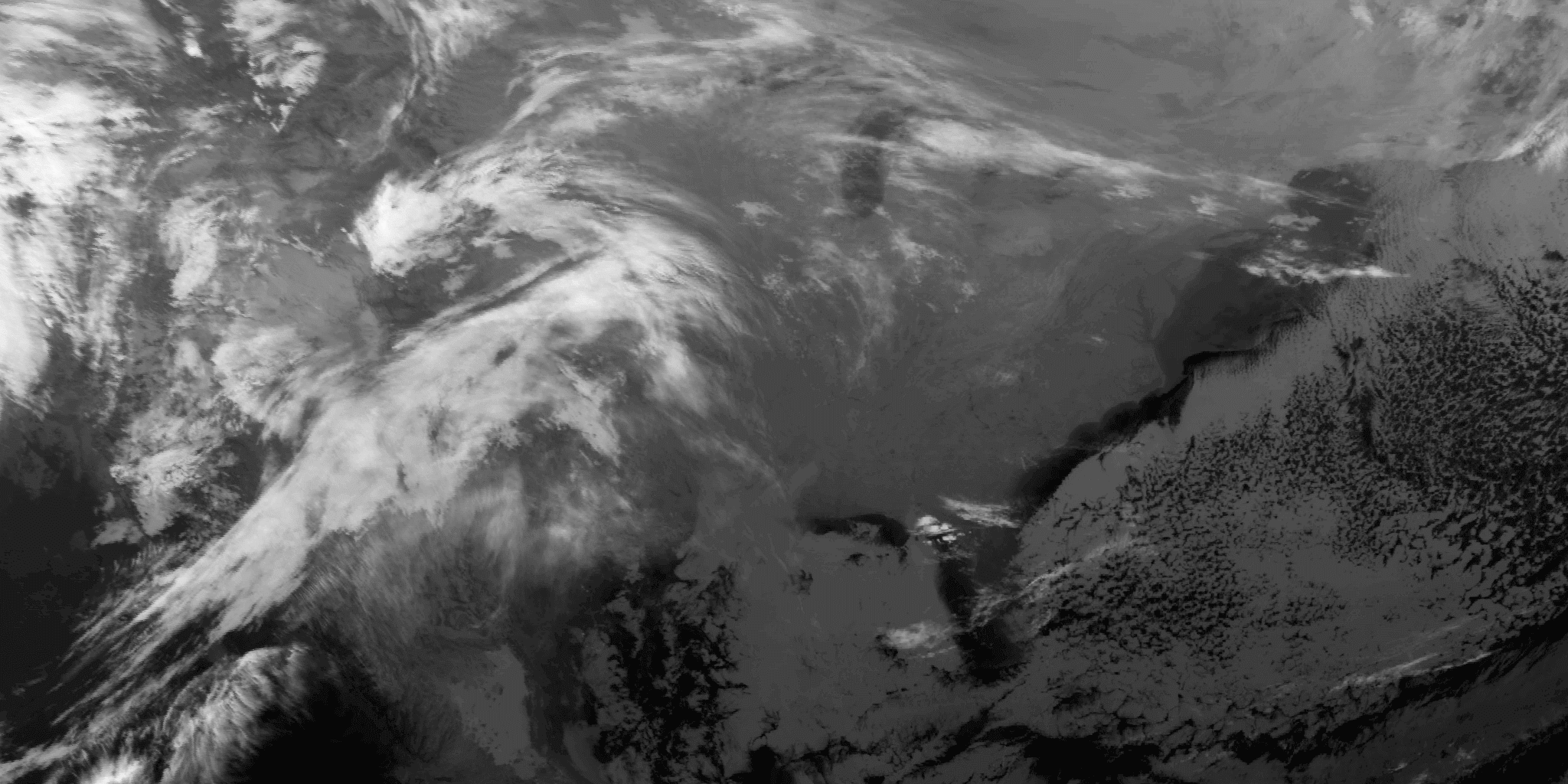}}\hspace{0.01in}
\subfloat {\includegraphics[width=0.09\textwidth,trim=150 160 1036 0,clip]{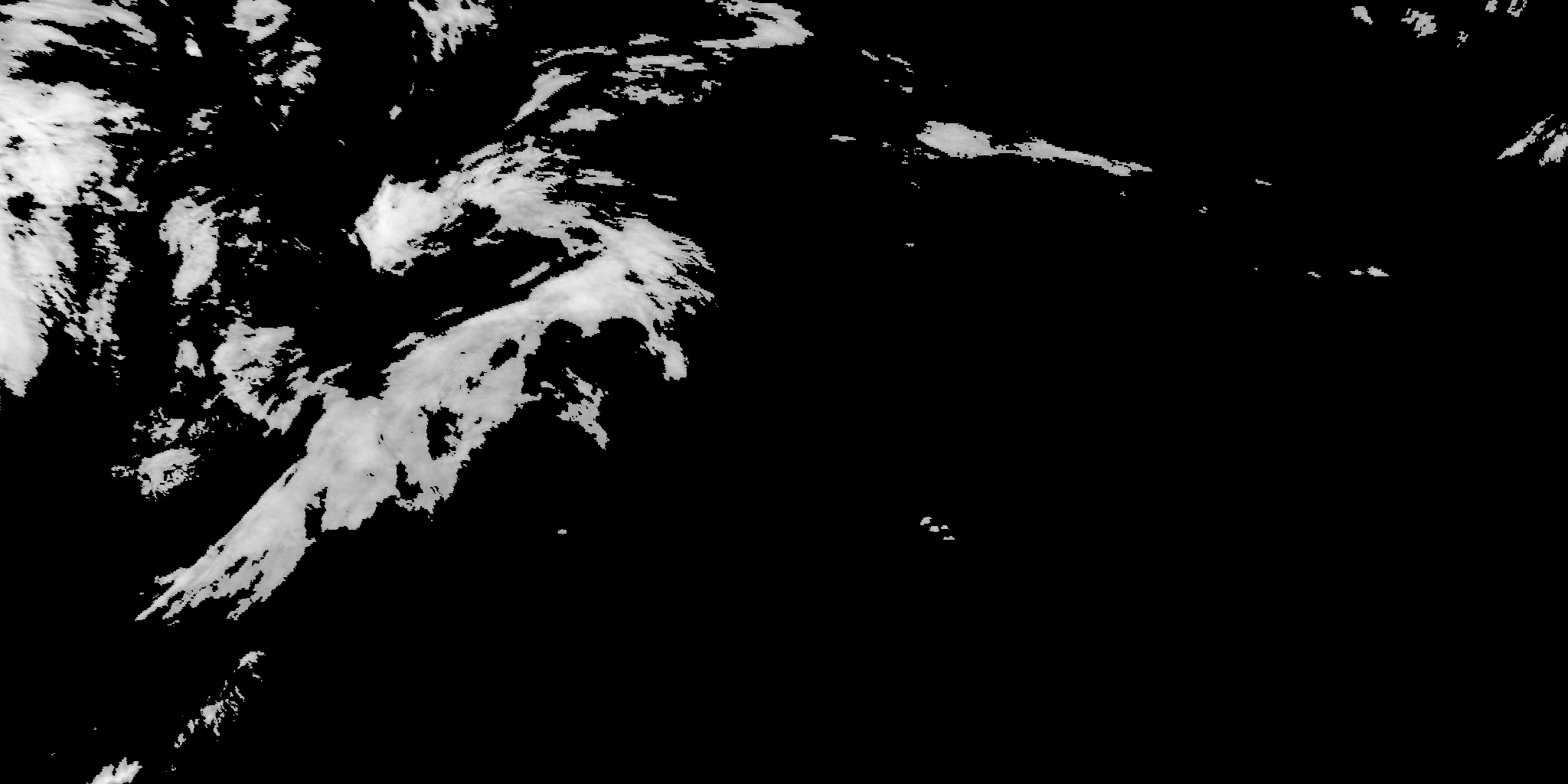}}\hspace{0.01in}
\subfloat {\includegraphics[width=0.09\textwidth,trim=150 160 1036 0,clip]{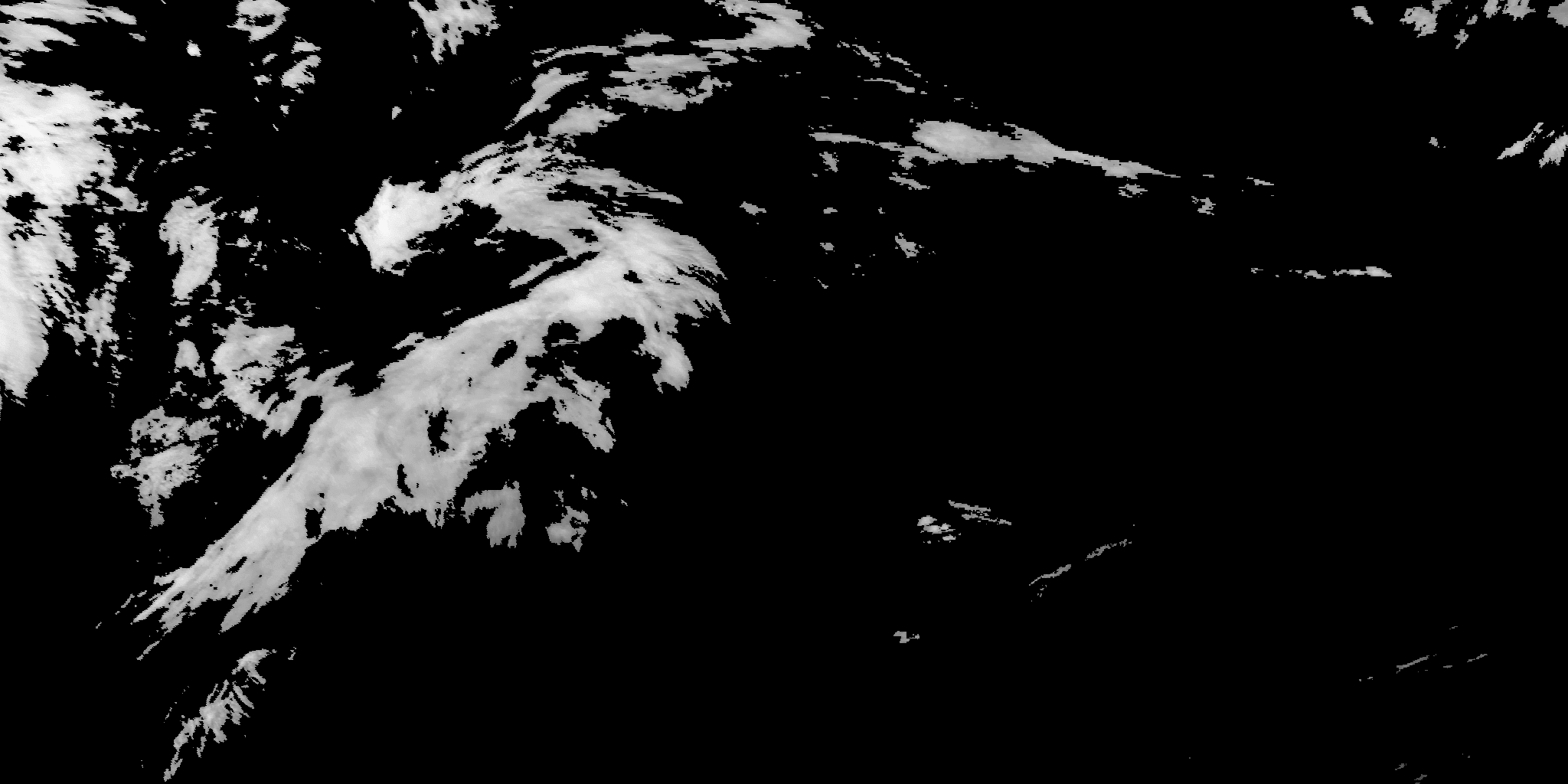}}\hspace{0.01in}
\subfloat {\includegraphics[width=0.09\textwidth,trim=150 160 1036 0,clip]{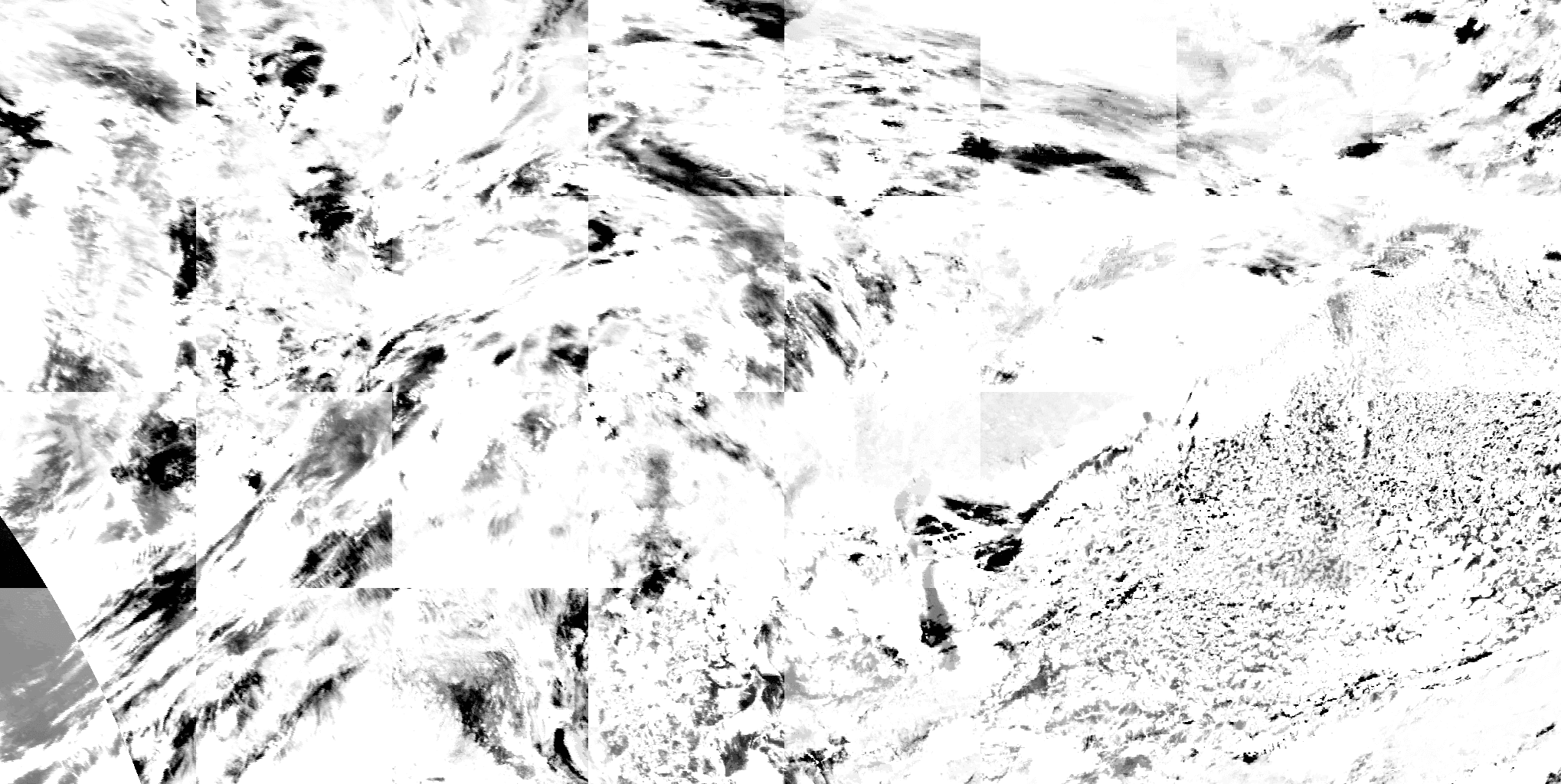}}\hspace{0.01in}
\subfloat {\includegraphics[width=0.09\textwidth,trim=150 160 1036 0,clip]{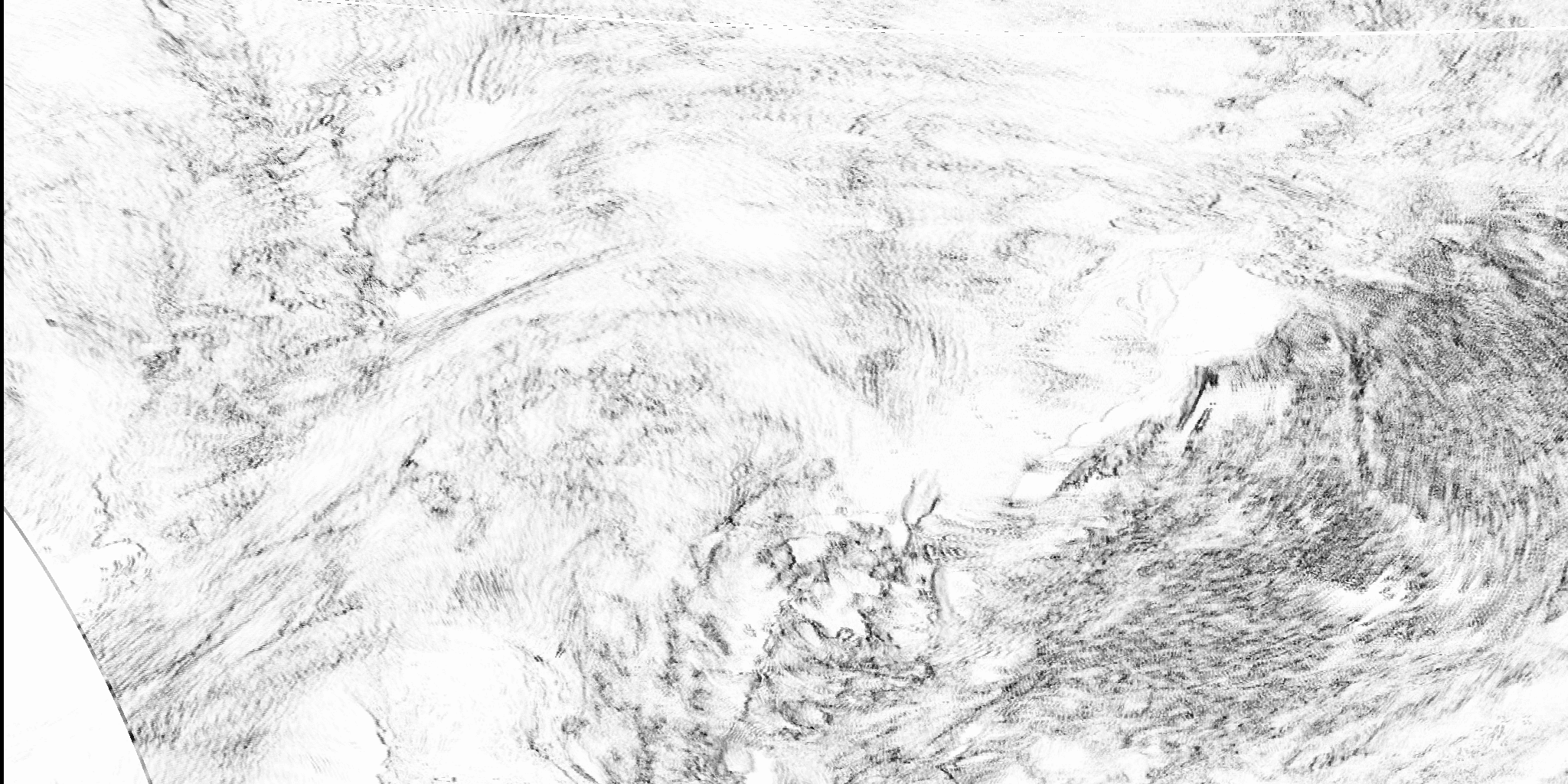}}

\vspace{-0.1in}

\subfloat {\includegraphics[width=0.09\textwidth,trim=250 400 1400 250,clip]{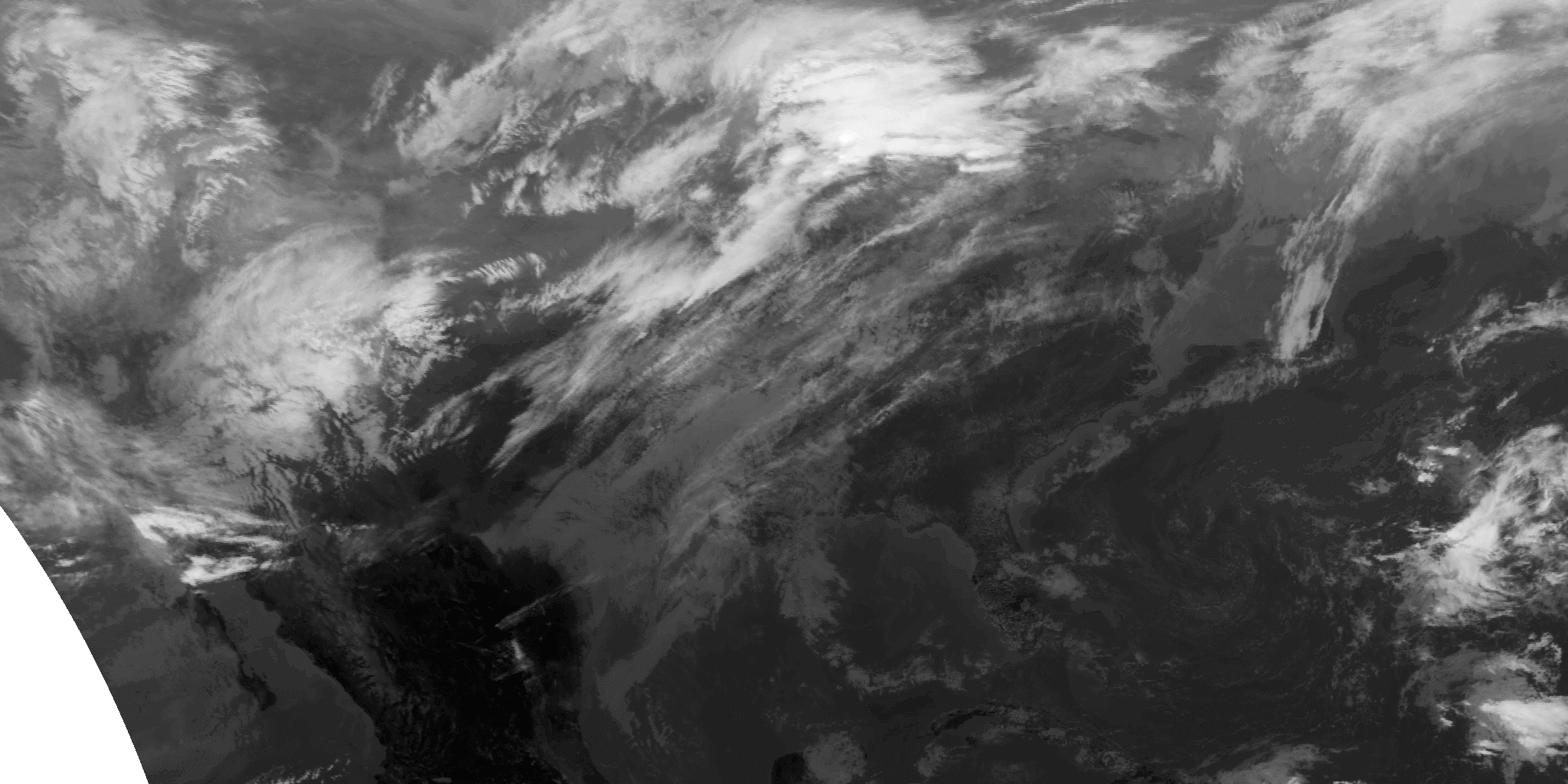}}\hspace{0.01in}
\subfloat {\includegraphics[width=0.09\textwidth,trim=250 400 1400 250,clip]{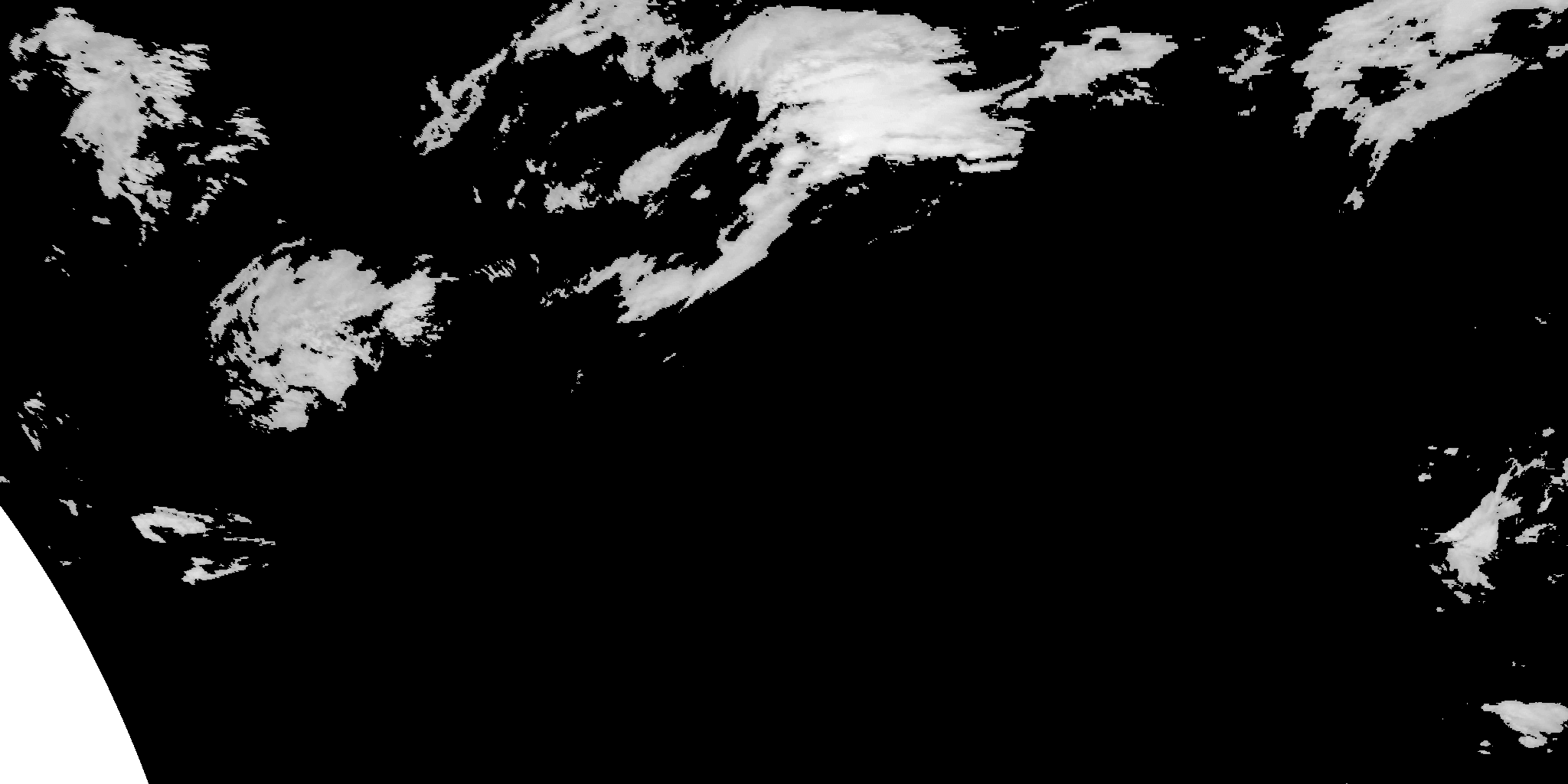}}\hspace{0.01in}
\subfloat {\includegraphics[width=0.09\textwidth,trim=250 400 1400 250,clip]{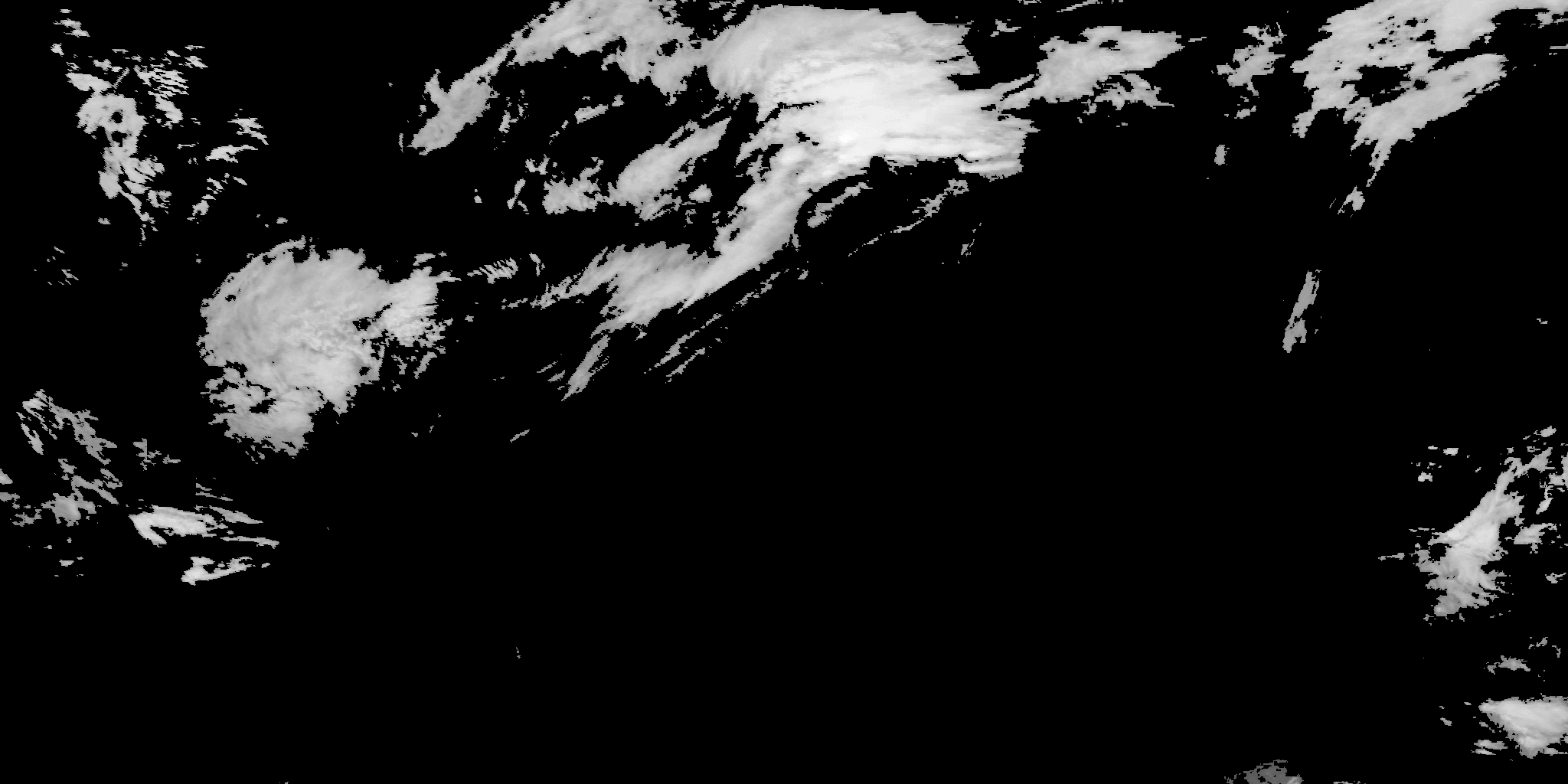}}\hspace{0.01in}
\subfloat {\includegraphics[width=0.09\textwidth,trim=250 400 1400 250,clip]{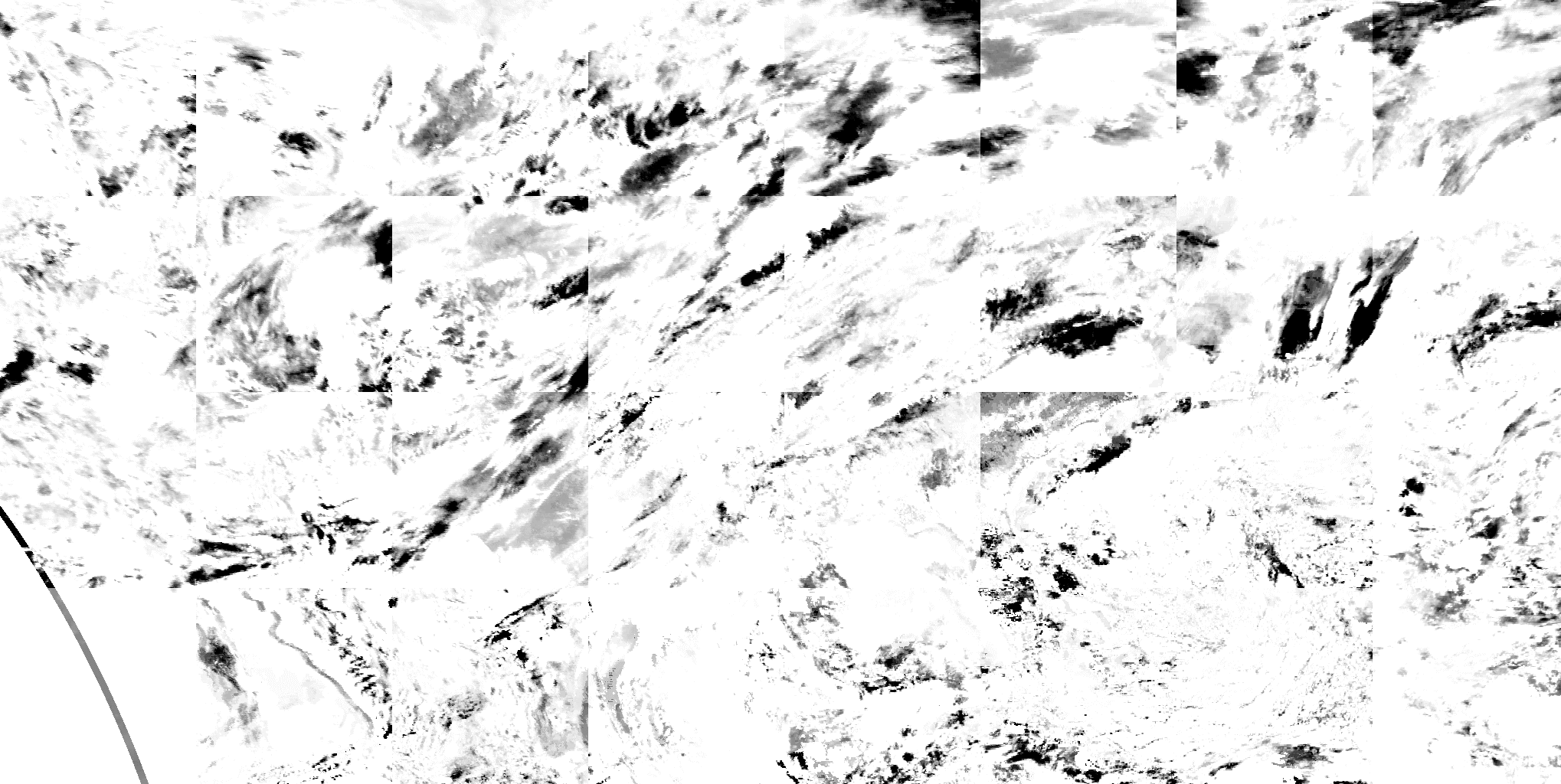}}\hspace{0.01in}
\subfloat {\includegraphics[width=0.09\textwidth,trim=250 400 1400 250,clip]{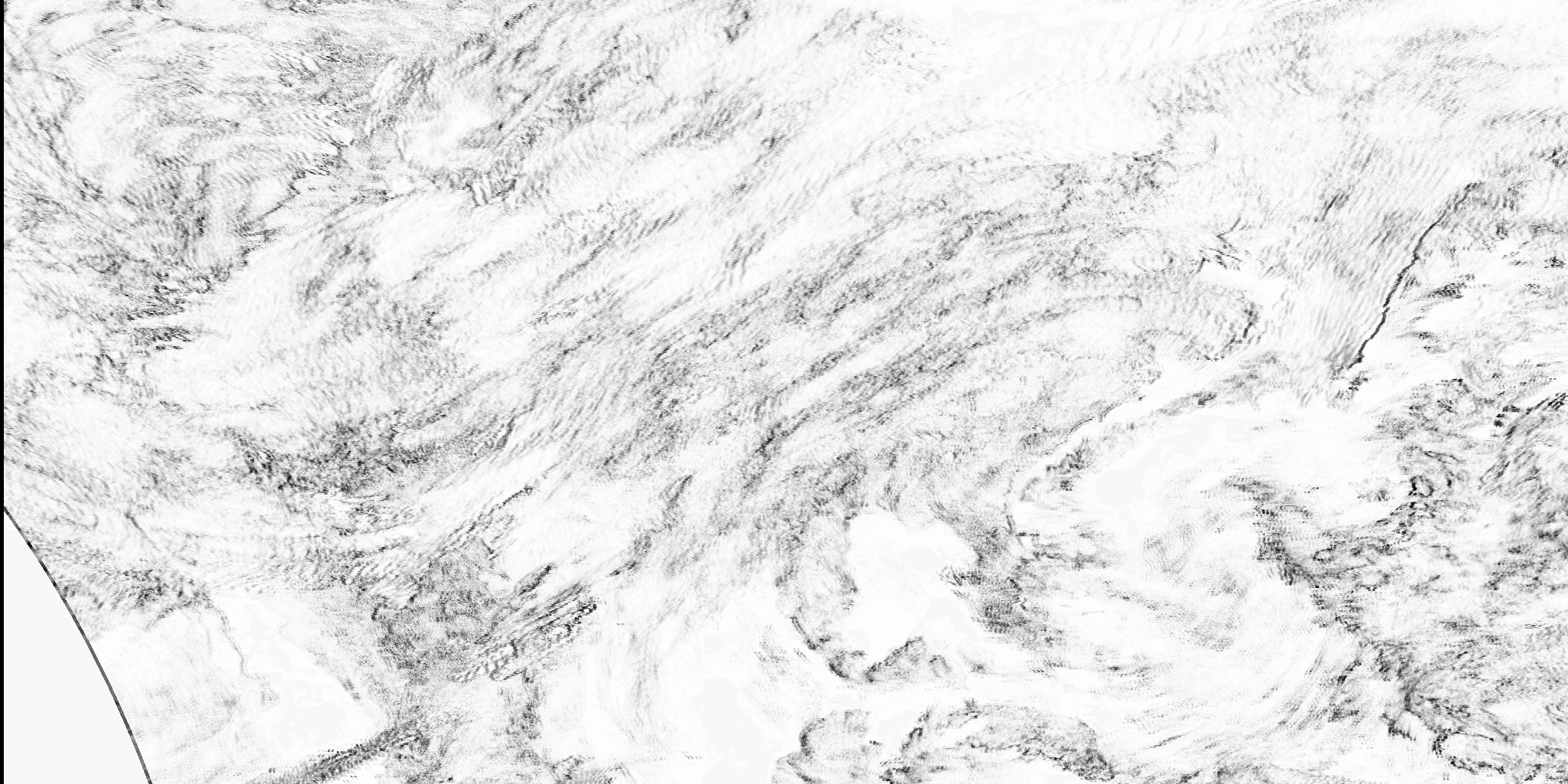}}

\vspace{-0.1in}

\subfloat[(a)] {\includegraphics[width=0.09\textwidth,trim=150 160 1036 0,clip]{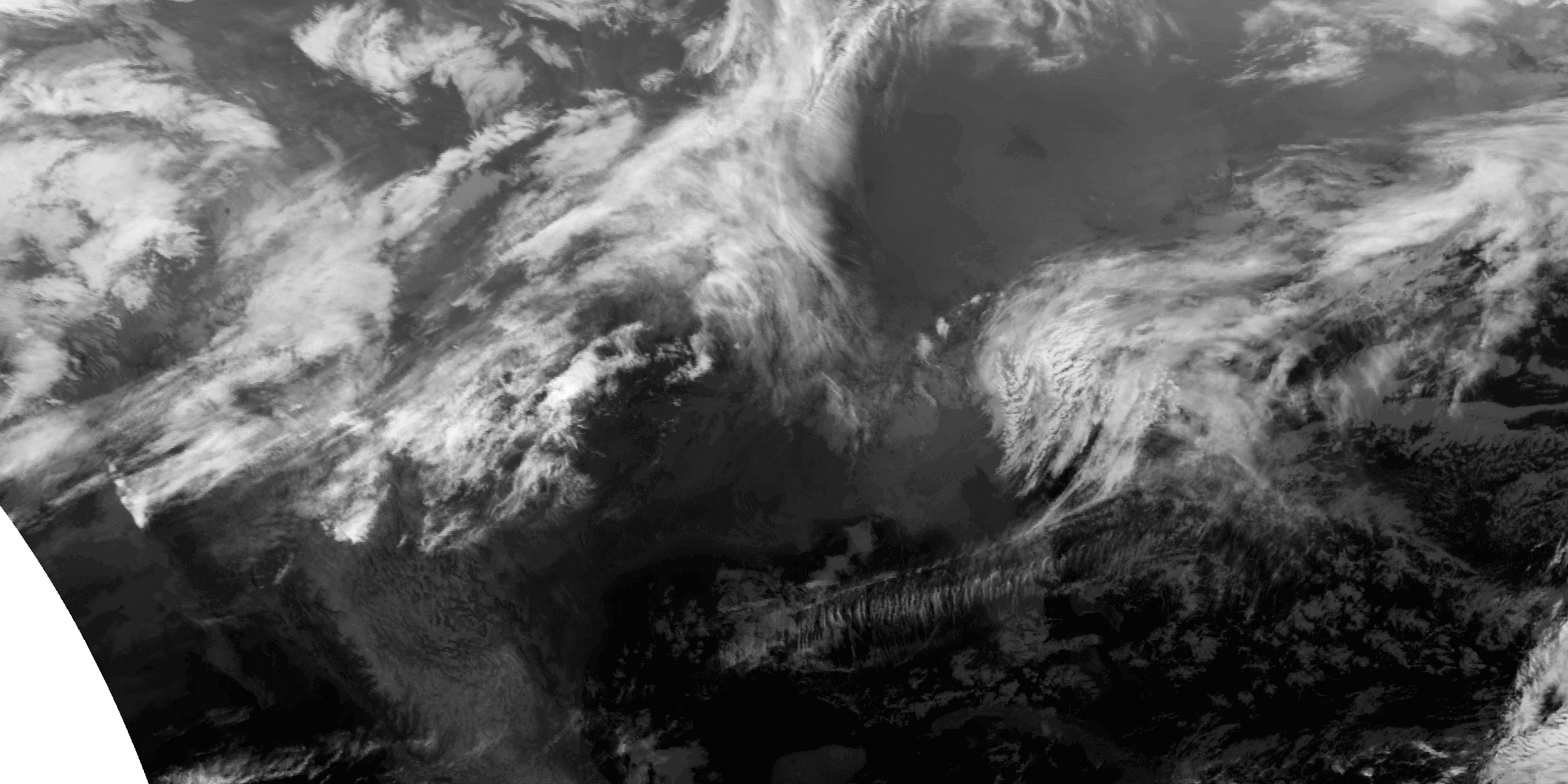}}\hspace{0.01in}
\subfloat[(b)] {\includegraphics[width=0.09\textwidth,trim=150 160 1036 0,clip]{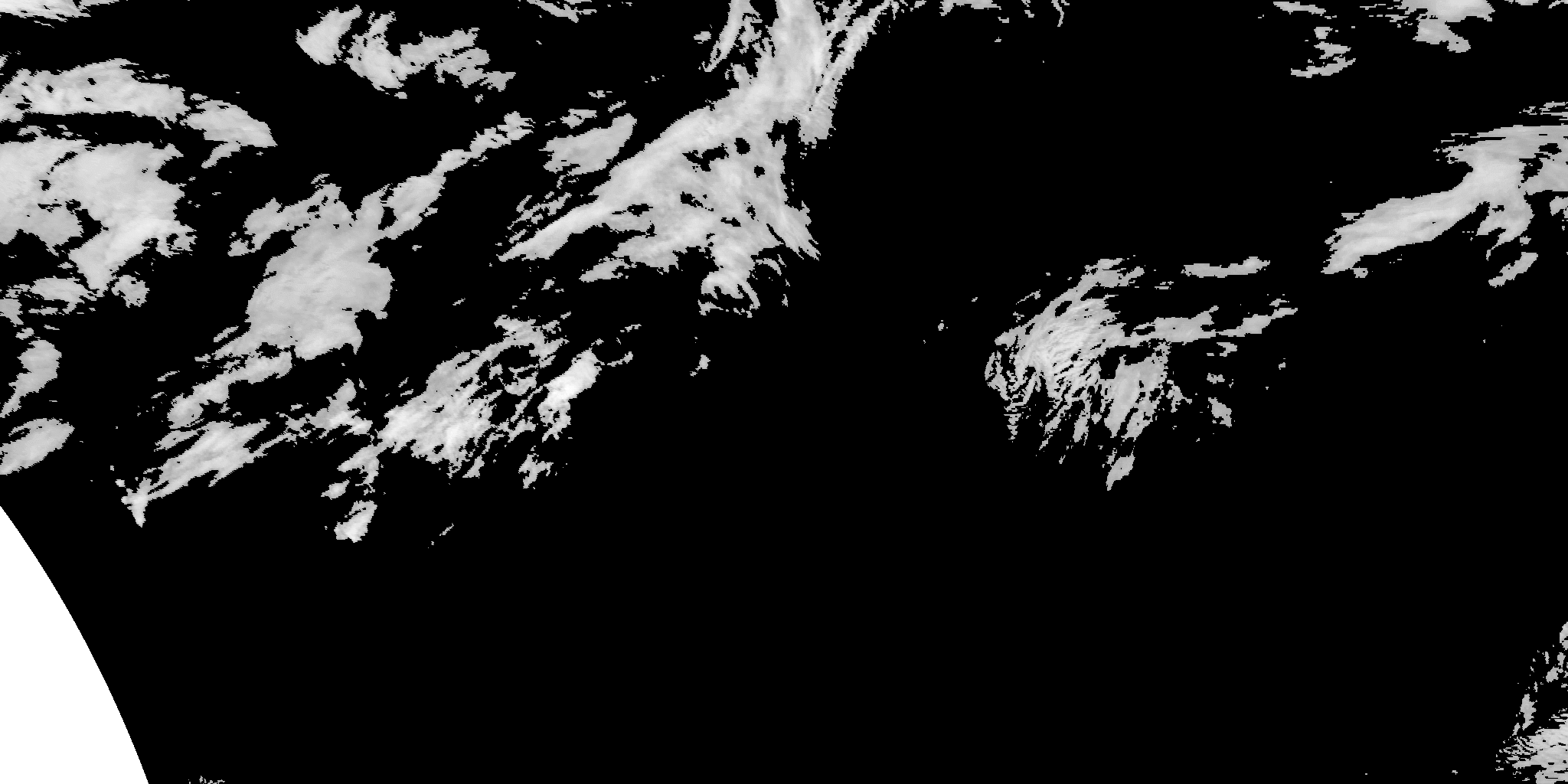}}\hspace{0.01in}
\subfloat[(c)] {\includegraphics[width=0.09\textwidth,trim=150 160 1036 0,clip]{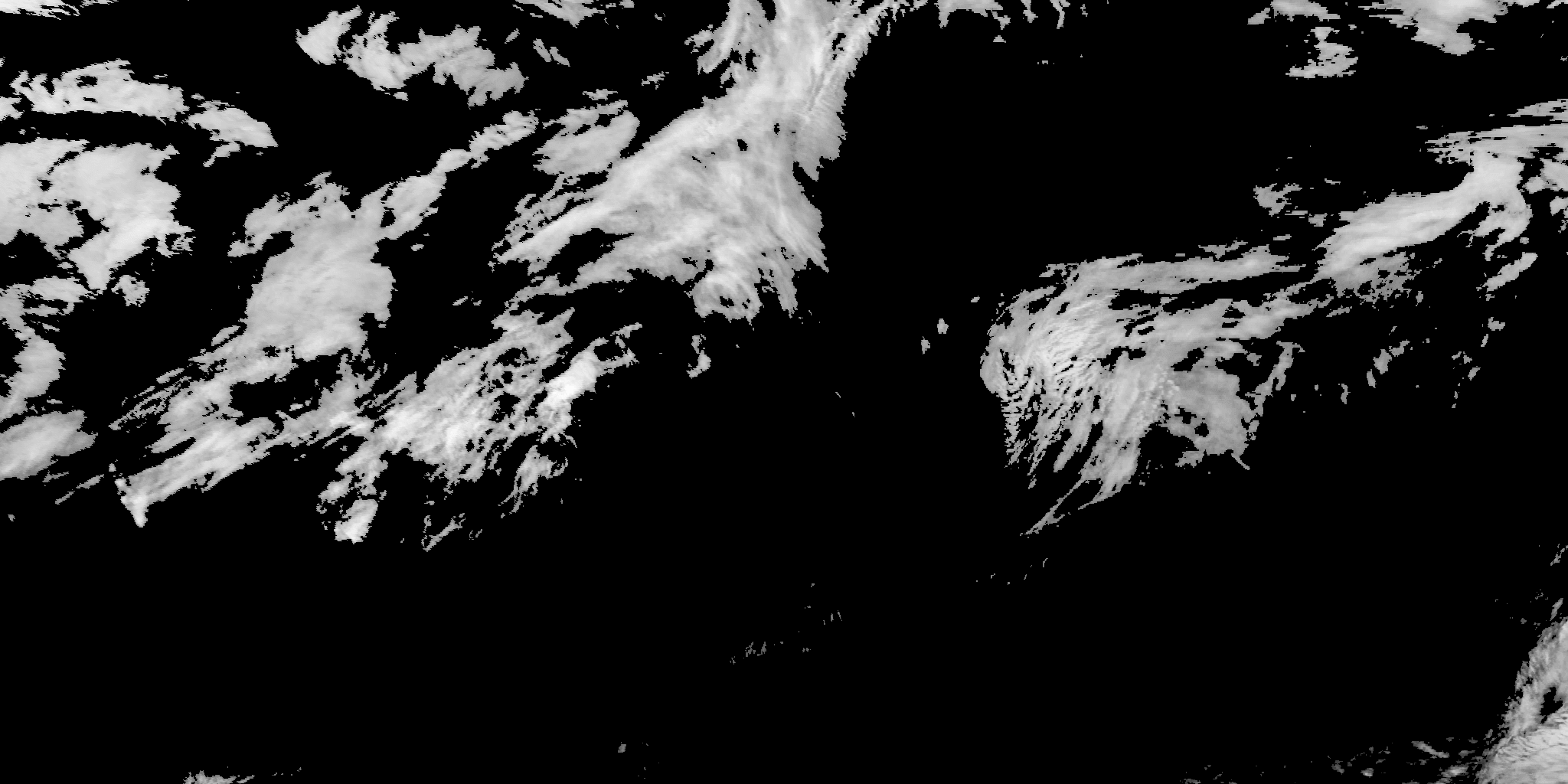}}\hspace{0.01in}
\subfloat[(d)] {\includegraphics[width=0.09\textwidth,trim=150 160 1036 0,clip]{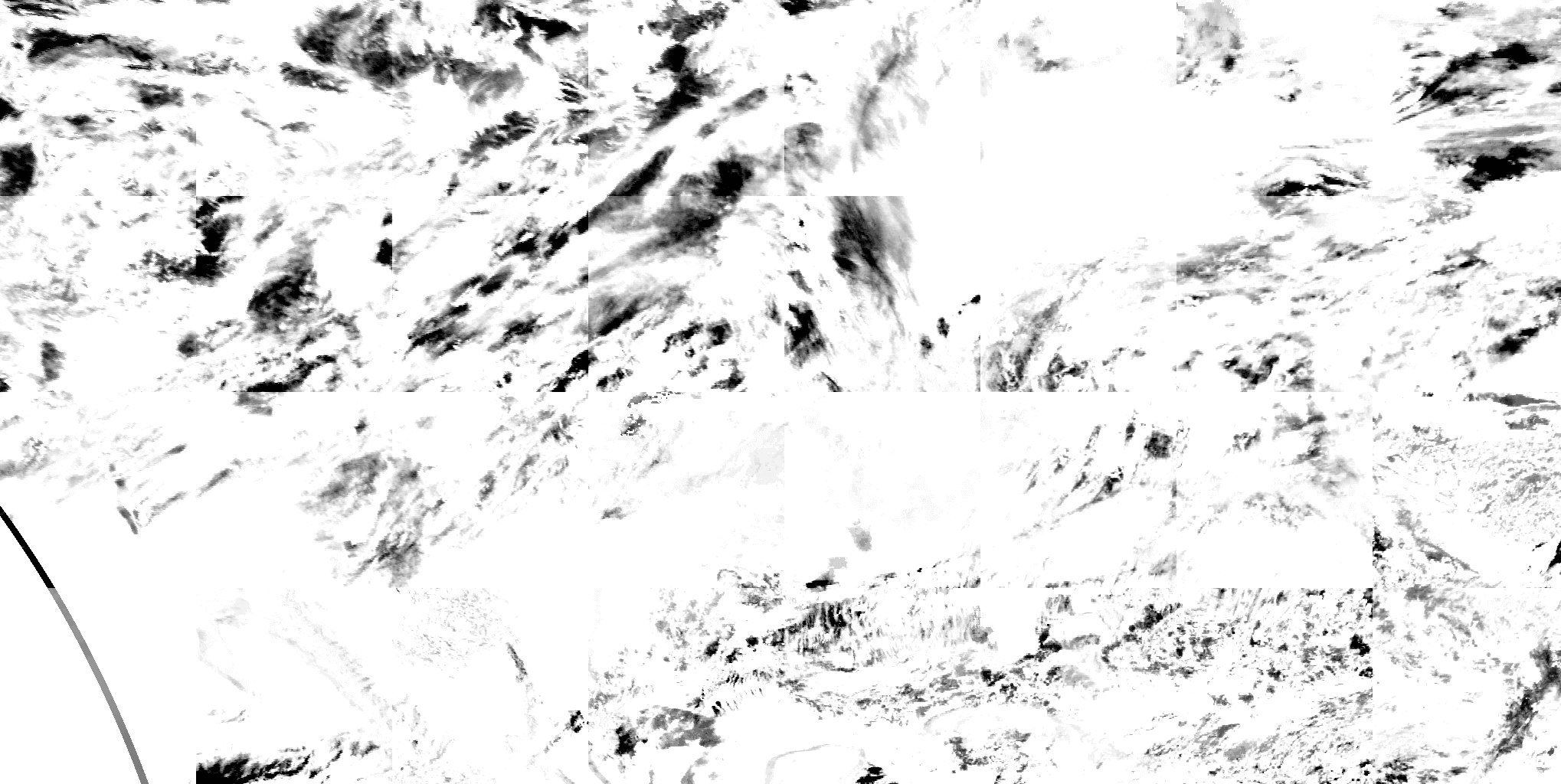}}\hspace{0.01in}
\subfloat[(e)] {\includegraphics[width=0.09\textwidth,trim=150 160 1036 0,clip]{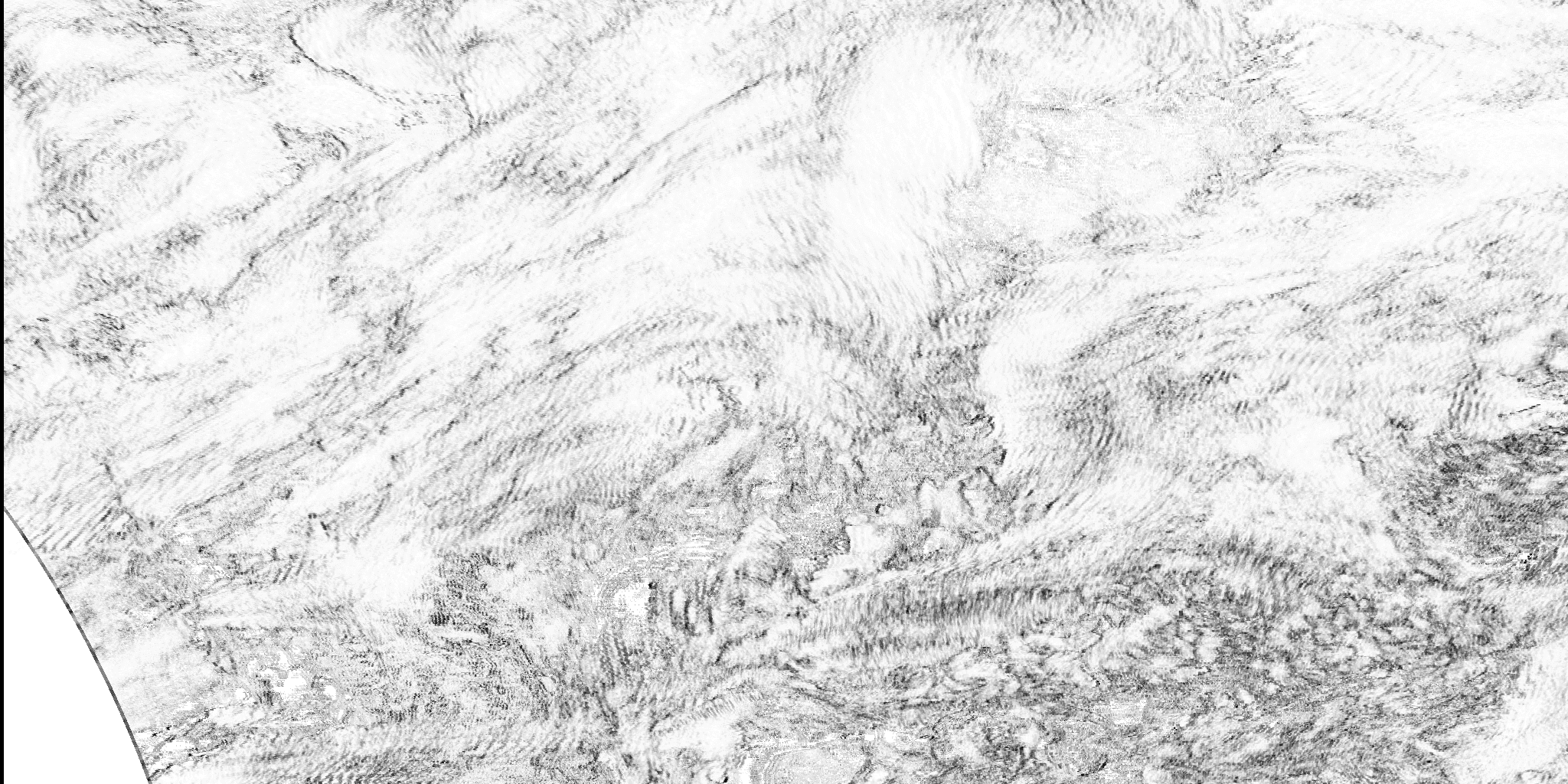}}
\caption{Cropped satellite images. (a) The original data. (b) Segmented high clouds with single threshold. (c) Segmented high clouds with GMM. (d) Cross-correlation in~\cite{leese1970determination}. (e) Correlation with motion prior.}
\label{fig:compare}
\end{figure}

We can conclude that our motion correlation is \textit{temporally smoothed} and the cross-correlation is \textit{spatially smoothed} by comparing Eq.~\eqref{eq:motion_corr} and~\eqref{eq:motion_ref}. The cross-correlation feature focuses on the differences in only two images, and then it takes the average on a spatial range. On the other hand, our correlation feature, with motion prior, interprets movement accumulation during the entire time span. 
We further re-normalize both $M(\cdot, \cdot)$ and $M_0(\cdot, \cdot)$ to [0, 255] and visualize these two motion descriptors in the fourth and fifth columns of Fig.~\ref{fig:compare}, where we fix $\mathbf h$ to be 10 pixels, $T$ to be 5 hours, and $h_0$ to be 128 pixels. The cross-correlation feature (fourth column of Fig.~\ref{fig:compare}) is noncontinuous across the area boundary. In image time series, the cross-correlation feature expresses less consistent positive/negative correlation in one neighborhood than our motion correlation does. Compared with the cross-correlation feature, our motion correlation feature (fifth column of Fig.~\ref{fig:compare}) shows more consistent texture with the cloud motion direction.

\subsection{Data Partition}\label{sec:data_par}

In this section, we use the widely-used ``sliding windows'' in~\cite{papageorgiou2000trainable} as the first-step detection. Sliding windows with an image pyramid help us capture the comma-shaped clouds at various scales and locations. Because most comma-shaped clouds are in the high sky, we run our sliding windows on the segmented cloud images. We set 21 dense $L\times L$ sliding windows, where $L\in \left\{128, 128 \cdot 8^{1/20}, \cdots, 128 \cdot 8^{19/20}, 1024 \right\}$. For each sliding window size $L$, the movement pace of the sliding window is $\left \lfloor L/8 \right \rfloor$, where $\left \lfloor \cdot \right \rfloor$ is the floor function. Under that setting, each satellite image has more than $10^{4}$ sliding windows, which is enough to cover the comma-shaped clouds in different scales.

Before we apply machine learning techniques, it is important to define whether a given bounding box is positive or negative. Here we use the Intersection over Union metric (IoU)~\cite{pascal-voc-2012} to define the positive and negative samples, which is also a common criterion in object detection. We set bounding boxes with IoU greater than a value to be the positive examples, and those with IoU = 0 to be the negative samples.

\begin{figure}[tbp!]\centering
\includegraphics[width=0.45\textwidth,trim=1cm 2.3cm 2.5cm 3.6cm,clip]{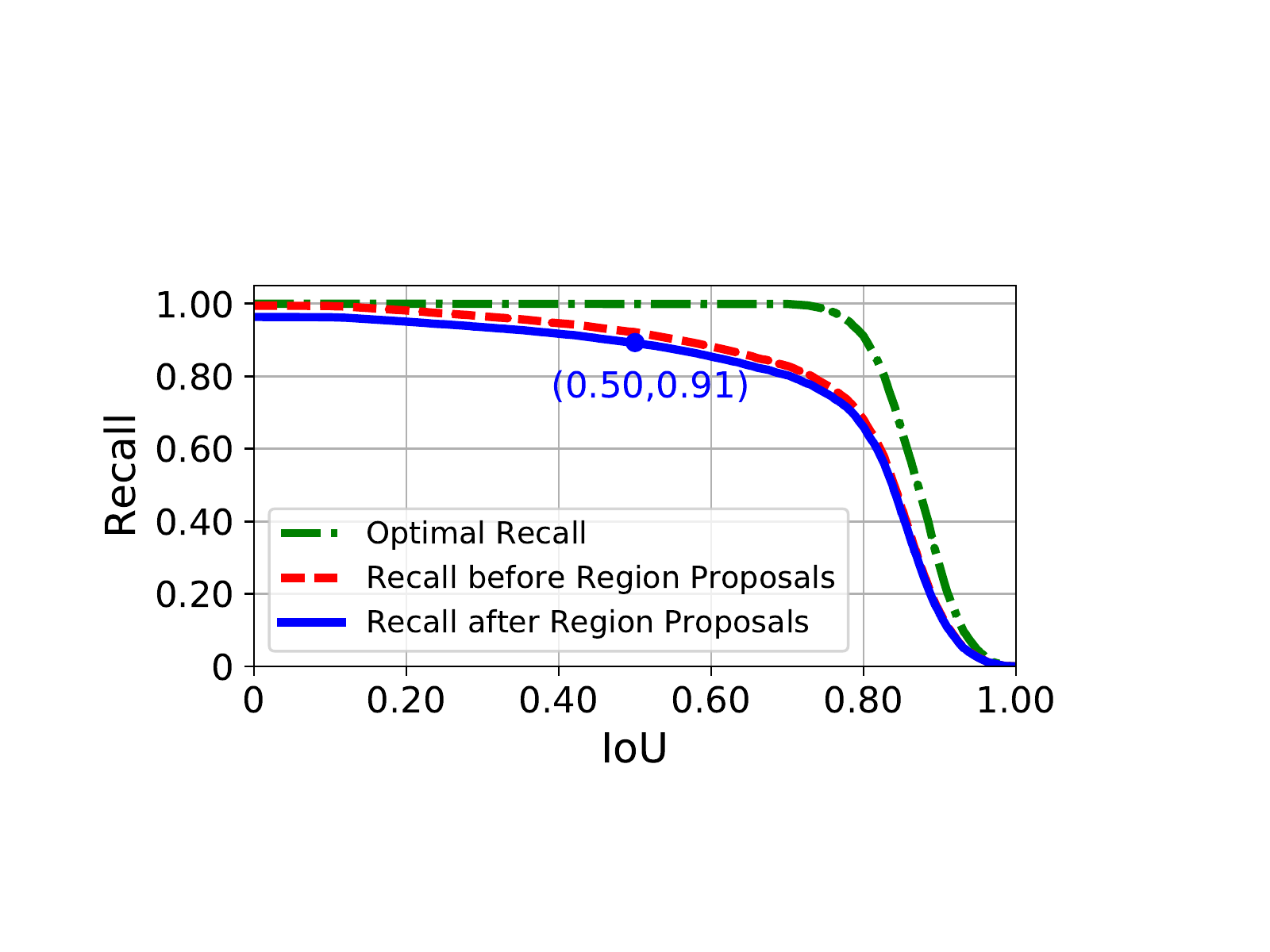}
\caption{IoU-Recall curve in the Region Proposal steps. The blue dot on the blue curve is our final IoU choice, with the corresponding recall of 0.91 .}
\label{fig:recall}
\end{figure}

A suitable IoU threshold should strike a balance between high recall and high accuracy of the selected comma-shaped clouds. Several factors prevent us from achieving a perfect recall rate. First, we only choose limited sizes of sliding windows with limited strides. Second, some of the satellite images are (partially) corrupted and unsuitable for a data-driven approach. Third, some cloud patches are in a lower altitude, hence they are removed in the high-cloud segmentation process in Sec.~\ref{sec:segmentation}. Fourth, we design simple classifiers to filter out most sliding windows without comma-shaped clouds (see Sec.~\ref{sec:regions}). Though we can get high efficiency by region proposals, the method inadvertently filters a small portion of true comma-shaped clouds. We show the IoU-recall curves in Fig.~\ref{fig:recall} for analyzing the effect of these factors to the recall rate. We provide our choice of IoU=0.50 as the blue dot in the plot and explain the reasons below.

Among the three curves in Fig.~\ref{fig:recall}, the green curve, marked as the \textit{Optimal Recall}, indicates the theoretical highest recall rate we can obtain with IoU changes. Because we have strong requirements to the sizes and locations of sliding windows in our algorithm, but do not apply those restrictions to human labelers, labeled regions and sliding windows cannot have a 100\% overlap due to human perception variations. Thus, we use the maximum IoU between each labeled region and all sliding windows as the highest theoretical IoU of this algorithm.
The red curve, marked as \textit{Recall before Region Proposals}, indicates the true recall we can get which considers missing images, image corruption, and high-cloud segmentation errors. Within our dataset, there are 11.26\% (5,926) of satellite images that are missing from the NOAA satellite image dataset, 0.36\% (188) that no recognized clouds, and 3.33\% (1,751) that have abnormally low contrast. Though low contrast level or dark images can be adjusted by histogram equalization, the pixel brightness values do not completely follow the GMMs estimated in the background extraction step.
Some high clouds are mistakenly removed with the background. In that experimental setting, this curve is the highest recall we can get before region proposals. 
The blue curve, marked as \textit{Recall after Region Proposals}, indicates the true recall we can get after region proposals, where the detailed process to design region proposals is in the following Sec.~\ref{sec:regions}.

The positive training samples consist of sliding windows whose IoU with labeled regions are higher than a carefully chosen threshold in order to guarantee both a reasonably high recall and a high accuracy. As a convention in object detection tasks, we expect IoU threshold $\geq$ 0.50 to ensure visual similarity with manually labeled comma-shaped clouds, and a reasonably high recall rate ($\geq 90\%$) in total for enough training samples. Finally, the IoU threshold is set to be 0.50 for our task. The recall rate is 92.26\% before region proposals and 90.66\% after region proposals. 

After establishing these boundaries, we partition the dataset into three parts: \textit{training set}, \textit{cross-validation set}, and \textit{testing set}. We use the data of the first 250 days of the year 2008 as the training set, the last 116 days of that year as the cross-validation set, and data from the years 2011 and 2012 as the testing set. 
The separation of the training set is due to the unusually large number of severe storms in 2008. The storm distribution ratio in the training, cross-validation, and testing sets are roughly 50\% : 15\% : 35\%. There are strong data dependencies between consecutive images. Splitting our data by time rather than randomly breaks this type of dependencies and more realistically emulates the scenarios within our system. This data partitioning scheme is also valid for the region proposals described in Sec.~\ref{sec:regions}.

\subsection{Region Proposals}\label{sec:regions}

In this stage, we design simple classifiers to filter out a majority of negative sliding windows. This method was also applied in~\cite{dalal2005histograms}. Because only a very small proportion of sliding windows generated in Sec.~\ref{sec:data_par} contain comma-shaped clouds, we can save computation in subsequent training and testing processes by reducing the number of sliding windows.

We apply three weak classifiers to decrease the number of sliding windows. 
The first classifier removes candidate sliding windows if their average pixel intensity is out of the range of [50, 200]. Comma-shaped clouds have typical shape characteristics that the cloud body part consists of dense clouds, but the dry tongue part is cloudless. Hence, the average intensity of a well-cropped patch should be within a reasonable range, neither too bright nor too dark. Finally, this classifier removes most cloudless bounding boxes while keeping over 98\% of the positive samples.

The second classifier uses a linear margin to separate positive examples from negative ones. We train this linear classifier on all the positive sliding windows with an equal number of randomly chosen negative examples, and then validate on the cross-validation set. All the sliding windows are resized to $256 \times 256$ pixels and vectorized before feeding into training, and the response variable is positive (1) or negative (0). As a result, the classifier has an accuracy of over 95\% on the training set and over 80\% on the cross-validation set. To ensure a high recall of our detectors, we output probability of each sliding window and then set a low threshold value. Sliding windows that output probability less than this threshold value are filtered out The threshold ensures that no positive samples are filtered out. We randomly change the train-test split for ten rounds and set the final threshold to be 0.2.

Finally, we compute the pixel-wise correlation $\gamma$ of each sliding window $I$ with the average comma-shaped cloud $I_0$. This correlation captures the similarity to a comma shape. $\gamma$ is computed as:
\begin{equation}
\gamma = \frac{I \cdot I_0}{\left \| I \right \|_{L_2} \cdot \left \| I_0 \right \|_{L_2}}\;.
\label{eq:corr}
\end{equation}

\begin{figure}[tbp!]
\begin{minipage}{0.95\linewidth}
\centering
\includegraphics[width=\textwidth,trim=0.2cm 2.3cm 0.56cm 3.2cm,clip]{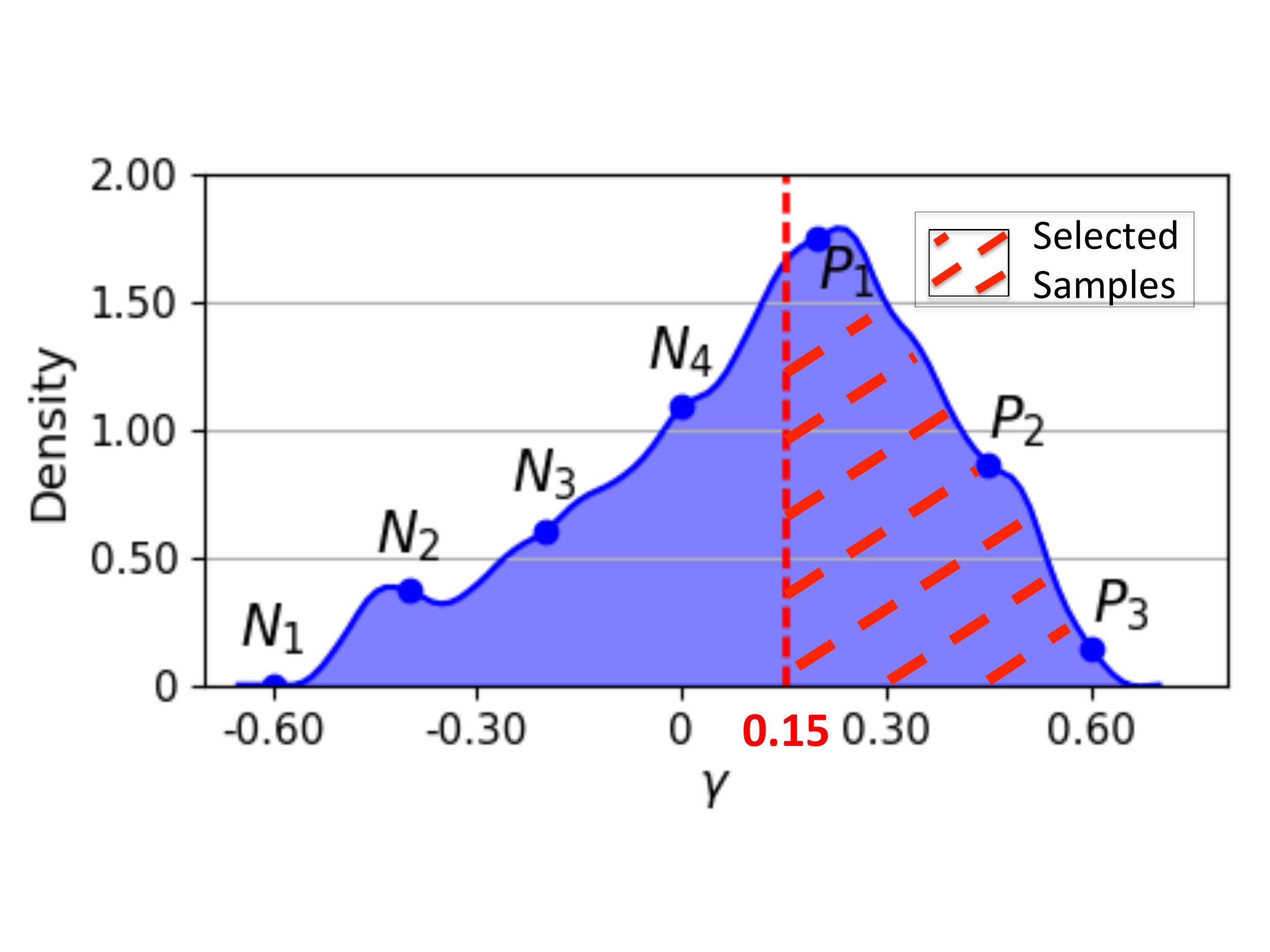}
\end{minipage}
\begin{minipage}{0.95\linewidth}
  \centering
  \begin{threeparttable}
  \begin{tabular}{ |m{0.14\linewidth}|m{0.15\linewidth}|m{0.15\linewidth}|m{0.15\linewidth}|m{0.15\linewidth}| }
    \hline
    Marker & $N_1$ & $N_2$ & $N_3$ & $N_4$ \\
    \hline
    $\gamma$ & -0.60 & -0.40 & -0.20 & 0\\
    \hline
    Example & 
      \includegraphics[scale=0.08]{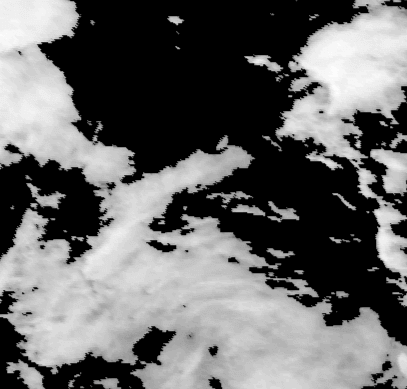} & 
      \includegraphics[width=\linewidth]{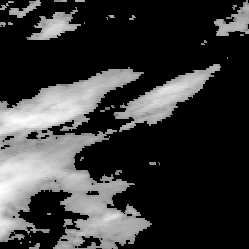} & 
      \includegraphics[width=\linewidth]{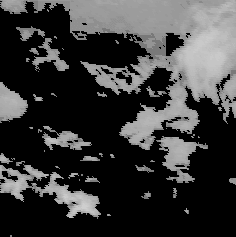} & 
      \includegraphics[width=\linewidth]{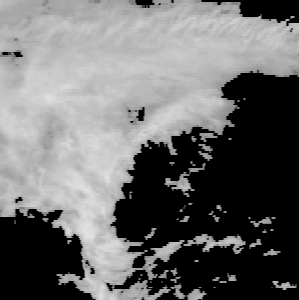}\\
    \hline\hline 
    Marker & $P_1$ & $P_2$ & $P_3$ & --\\ 
    \hline
    $\gamma$ & 0.20 & 0.40 & 0.60 &Avg.\\ 
    \hline
    Example & 
      \includegraphics[width=\linewidth]{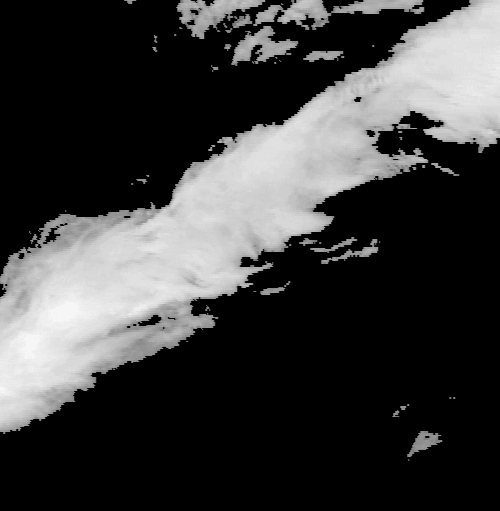} & 
      \includegraphics[width=\linewidth]{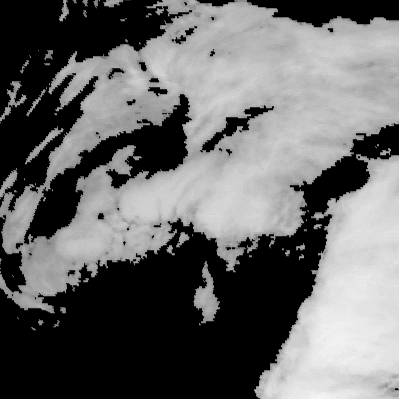} & 
      \includegraphics[width=\linewidth]{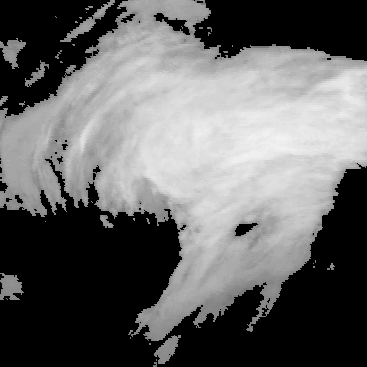} & 
      \includegraphics[width=\linewidth]{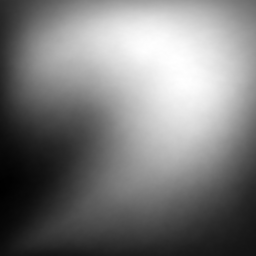}\\
    \hline \hline 
    TS$^{*}$& TS & Lightning & Hail & Marine\\ 
    Category& Wind & & & TS Wind\\ 
    \hline
    Avg. Image& 
      \includegraphics[width=\linewidth]{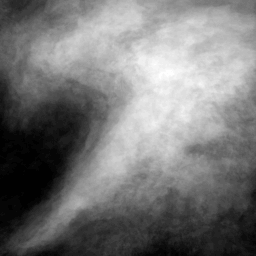} & 
      \includegraphics[width=\linewidth]{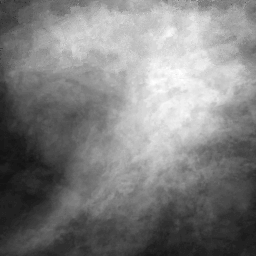} & 
      \includegraphics[width=\linewidth]{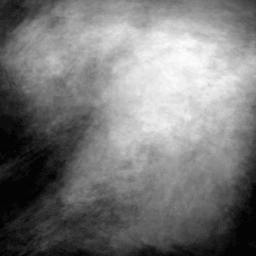} & 
      \includegraphics[width=\linewidth]{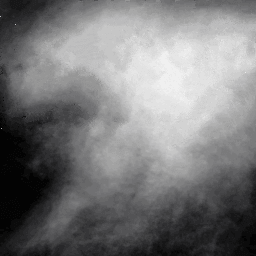}\\
    \hline
  \end{tabular}
  \begin{tablenotes}\footnotesize
\item[*] TS: Thunderstorm.
\end{tablenotes}
\end{threeparttable}
\end{minipage}
\caption{\textit{Top}: The correlation probability distribution of all sliding windows. \textit{Middle}: 
Some segmented image examples. The last example image is the average image of the manually labeled regions in the training set. The correlation score $\gamma$ is defined in Eq.~\eqref{eq:corr}, and the diagram is the normalized probability distribution of $\gamma$ of the training set.
\textit{Bottom}: Average comma-shaped clouds in different categories. } 
\label{fig:distribute}
\end{figure}

Because there are no visual differences between different categories of storms (as shown in the last row of the table in Fig.~\ref{fig:distribute}), $I_0$ is the average labeled comma-shaped clouds in the training set. The computation process of $I_0$ consists of the following steps. First, we take all the labeled comma-shaped clouds bounding boxes in the training set and resize them to $256 \times 256$. Next, we segment the high-cloud part from each image using the method in Sec.~\ref{sec:segmentation}. Finally, we take the average of the high-cloud parts. The resulting $I_0$ is marked as Avg. in the middle row of the table in Fig.~\ref{fig:distribute}. To be consistent in dimensions, every sliding window $I$ is also resized to $256 \times 256$ in Eq.~\eqref{eq:corr}.

The higher correlation $\gamma$ indicates that a cloud patch has the appearance of comma-shaped clouds. Based on this observation, a simple classifier is designed to select sliding windows whose $\gamma$ is higher than a certain threshold.
Fig.~\ref{fig:distribute} serves as a reference to choose a customized threshold of $\gamma$. The distribution and some example images of $\gamma$ is listed in the table of Fig.~\ref{fig:distribute}. In the training and cross-validation sets, less than 1\% of positive examples have a $\gamma$ value lower than 0.15. So, we use $\gamma \geq 0.15$ as the final filter choice to eliminate sliding window candidates. 

The region proposal process only permits about $10^{3}$ bounding boxes per image, which is only one tenth of the initial number of bounding boxes. As shown in Fig.~\ref{fig:recall}, the region proposals process does not significantly affect the recall rate, but it can save much time for the later training process.

\subsection{Construction of Weak Classifiers}~\label{sec:weakclassifier}

We design two sets of descriptive features to distinguish comma-shaped clouds. The first is the histogram of oriented gradients (HOG)~\cite{dalal2005histograms} feature based on segmented high clouds. For each of the region proposals, we compute the HOG feature of the bounding box. Because we compute the HOG feature on the segmented high clouds, we refer to it as \textit{Segmented HOG} feature in the following paragraphs. The second is the histogram feature of each image crop based on the motion prior image, where the texture of the image reflects the motion information of cloud patches. We fine-tune the parameters and show the accuracy on cross-validation set in Table~\ref{tb:HOGLBP}. We use Segmented HOG setting~\#4 and Motion Histogram Setting~\#5 as the final parameter setting in our experiments because it has a better performance on the cross-validation set. The feature dimension is 324 for Segmented HOG and 27 for Motion Histogram.

\begin{table}[tb]
\begin{center}
\caption{Avg. Accuracy of weak classifiers for the segmented HOG and motion histogram features in different parameter settings}
\begin{threeparttable}
\begin{tabular}{c|c|c|c|c}
  \hline
Seg. HOG & Orientation & Pixels/ & Cells/ & (\%)Avg. \\
Settings & Directions & Cell &  Block & Accuracy \\
  \hline
\#1 & 9 & $64 \times 64 $ & $1\times 1$ & $69.88 \pm 1.15 $ \\
\#2 & 18 & $64 \times 64 $ & $1\times 1$ & $ 70.65 \pm 1.25 $ \\
\#3 & 9 & $128 \times 128 $ & $1\times 1$ & $ 61.21 \pm 0.62 $ \\
\textbf{\#4} & \textbf{9} & \textbf{64} $\times$ \textbf{64} & \textbf{2} $\times$ \textbf{2} & \textbf{73.18} $\pm$ \textbf{0.98} \\
  \hline
  \hline
Motion Hist. & Pixels to & Time Span & Hist. & (\%)Avg. \\
  Settings & the West $\mathbf h^{*}$&in hours $T^{*}$&  Bins & Accuracy \\
  \hline
\#1 & 10 & 2 & 18 & $58.84 \pm 0.59$ \\
\#2 & 5 & 2 & 18 & $52.97 \pm 0.20$ \\
\#3 & 10 & 2 & 9 & $58.83 \pm 0.20$ \\
\#4 & 10 & 2 & 27 & $61.05 \pm 0.63$ \\
\textbf{\#5} & \textbf{10} & \textbf{5} & \textbf{27} & \textbf{63.25} $\pm$ \textbf{0.67} \\
  \hline
\end{tabular}
\begin{tablenotes}\footnotesize
\item[*] $\mathbf h$ and T have the same meaning as annotated in Eq.~\eqref{eq:motion_corr}.\\
\end{tablenotes}
\end{threeparttable}
\label{tb:HOGLBP}
\end{center}
\vspace{-0.2in}
\end{table}

\begin{figure}[tb!]\centering
\includegraphics[width=0.49\textwidth,trim=0cm 0cm 0cm 0cm,clip]{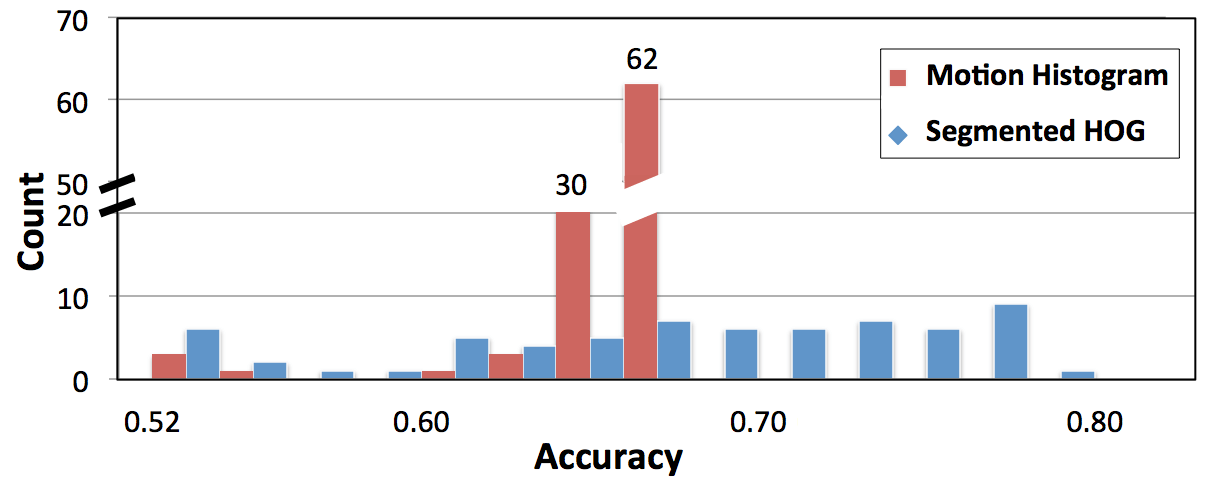}
\caption{The accuracy distribution of weak classifiers with Segmented HOG feature and Motion Histogram feature.}
\label{fig:weakclassifier}
\end{figure}

Since severe weather events have a low frequency of occurrence, positive examples only take up a very small proportion ($\sim 1 \%$) in the whole training set. To utilize negative samples fully in the training set, we construct 100 linear classifiers. Each of these classifiers is trained on the whole positive training set and an equal number of negative samples. We split and randomly select these 100 batches according to their time stamp so that their time stamps are not overlapped with other batches. We train each logistic regression model on the segmented HOG features and the motion histogram feature of the training set. Finally, we get 200 linear classifiers. We evaluate the accuracy of the trained linear classifiers based on a subset of testing examples whose positive/negative ratio is also 1-to-1. The average accuracy of the segmented HOG feature is 73.18\% and that of the motion histogram features is 63.25\%. The accuracy distribution of these two types of weak classifiers is shown in Fig.~\ref{fig:weakclassifier}. From the statistics and the figure, we know the Segmented HOG feature has a higher average accuracy than the Motion Histogram feature has, and a larger variation in the accuracy distribution. As shown in Fig.~\ref{fig:weakclassifier}, about 90\% of the classifiers on the motion histogram feature have an accuracy between 63\% and 65\%, while those on the segmented HOG feature distribute in a wider range from 53\% to 80\%. 

\subsection{AdaBoost Detector}~\label{sec:adaboost}

We apply the stacked generalization on the probability output of our weak classifiers~\cite{wolpert1992stacked}. For each proposed region, we use the probability output of the 200 weak classifiers as the input, and get one probability $p$ as the output. We define the proposed region is positive for $p \geq p_0$ or negative in other cases, where $p_0\in (0, 1)$ is our given cutoff value.

We adopt AdaBoost~\cite{freund1995desicion} as the method for stacked generalization because it achieves the highest accuracy on the balanced cross-validation set, as shown in Table~\ref{tb:ensemble}. All these classifiers are constructed on the training set and fine-tuned on the cross-validation set. Table~\ref{tb:Adaboost} shows the accuracy of the AdaBoost classifier with different parameters. 
For each set of parameters, we provide a 95\% confidence interval computed on 100 random seeds in both Table~\ref{tb:ensemble} and~\ref{tb:Adaboost}.
The classification accuracy reaches the maximum at 86.47\% with 40 leaf nodes and one layer. The AdaBoost classifier running on region proposals is our proposed comma-shaped cloud detector.

\begin{table}[tb]
\begin{minipage}{.49\linewidth}
\begin{center}
\caption{Accuracy of different stacked generalization methods on the cross-validation set}
\begin{threeparttable}
\begin{tabular}{c|c}
  \hline
    Method$^{*}$ & (\%)Accuracy \\
  \hline
  	LR & 85.10 $\pm$ 0.20\\
    Bagging & 81.98 $\pm$ 0.46\\
    RF & 82.34 $\pm$ 0.40 \\
    ERF & 82.45 $\pm$ 0.34\\
    GBM & 85.77 $\pm$ 0.25 \\
    \textbf{AdaBoost} & \textbf{86.47} $\pm$ \textbf{0.25}\\
  \hline
\end{tabular}
\label{tb:ensemble}
\begin{tablenotes}\footnotesize
\item[*] LR: Logistic Regression; RF: Random Forest; ERF: Extremely Random Forest; GBM: Gradient Boosting Machine with deviance loss.\\
\end{tablenotes}
\end{threeparttable}
\end{center}
\end{minipage}
\hfill
\begin{minipage}{.5\linewidth}
\begin{center}
\caption{Accuracy of the AdaBoost detector on the cross-validation set with different parameters}
\begin{tabular}{c|c|c}
  \hline
  Tree & Leaf  & (\%)Accuracy \\
   layer &  Nodes  & \\
  \hline
   & 20 & 85.49 $\pm$ 0.26\\
  \textbf{1} &  \textbf{40} &  \textbf{86.47} $\pm$ \textbf{0.25} \\
   & 60 & 86.45 $\pm$ 0.22 \\
   \hline
   & 20 & 86.11 $\pm$ 0.24 \\
  2 & 40 & 86.27 $\pm$ 0.24  \\
   & 60 & 84.97 $\pm$ 0.25  \\
  \hline
\end{tabular}
\label{tb:Adaboost}
\end{center}
\end{minipage}
\end{table}

We then run the AdaBoost detector on the testing set and then compute the ratio of the labeled comma-shaped clouds that our method can detect. For each image, we choose the detection regions that have the largest probability scores of having comma-shaped clouds (abbreviated as probability in this paragraph), and we ensure every two detection regions in one image have an IoU less than 0.30 --- a technique called non-maximum suppression (NMS) in object detection~\cite{rosten2006machine}. If one output region has IoU larger than 0.30 with another output region, we remove the one with lower probability from the AdaBoost detector. Finally, the detector outputs a set of sliding windows, with each region indicating one possible comma-shaped clouds.

In our experiment, the training set for ensembling is a combination of all 68,708 positive examples and a re-sampled subset of negative examples sized ten times larger than the size of positive examples ({\it i.e.}, 687,080). 
We carry out experiments with the Python 2.7 implementation on a server with the Intel$\textsuperscript{\textregistered}$ Xeon X5550 2.67GHz CPUs. We apply our algorithm on every satellite image in parallel.
If the cutoff threshold is set to be 0.50, the detection process for one image, from high-cloud segmentation to AdaBoost detector, costs about 40.59 seconds per image. Within that time, the high-cloud segmentation takes 4.69 seconds, region proposals take 14.28 seconds, and the AdaBoost detector takes 21.62 seconds. We only get two satellite images per hour now, and these three processes finish in a sequential order. If higher speed is needed, an implementation in C/C++ is expected to be substantially faster.

\section{Evaluation}\label{sec:result}

In this section, we present the evaluation results for our detection method. First, we present an ablation study. Second, we show that our method can effectively detect both comma-shaped clouds and severe thunderstorms. Finally, we compare our method with two other satellite-based storm detection schemes and show that our method outperforms both.

\subsection{Ablation Study}

\begin{table}[!tb]
\begin{center}
\caption{Accuracy of the AdaBoost classifier on the cross-validation set with different features}
\begin{threeparttable}
\begin{tabular}{c|c|c}
  \hline
    With high-cloud & Feature(s) & Accuracy (\%) \\
    segmentation & & \\
  \hline
   & HOG & 70.45 $\pm$ 1.13\\
  No & Motion Hist. & 55.84 $\pm$ 0.88\\
   & Combination & 80.30 $\pm$ 0.41 \\
  \hline
   & HOG & 74.01 $\pm$ 0.90 \\
 \textbf{Yes} & Motion Hist. & 65.32 $\pm$ 0.62\\
   & \textbf{Combination} & \textbf{86.47} $\pm$ \textbf{0.25}\\
   \hline
\end{tabular}
\label{tb:ablation}
\begin{tablenotes}\footnotesize
\item Here HOG with high-cloud segmentation = Segmented HOG feature; Motion Hist. = Motion Histogram Feature.
\end{tablenotes}
\end{threeparttable}
\end{center}
\end{table}

To examine how much each step contributes to the model, we carried out an ablation study and show the results in Table~\ref{tb:ablation}.
We enumerate all the combinations in terms of high-cloud segmentations and features. The first column indicates whether the region proposals are on the original satellite images or on the segmented ones. The second column separates HOG feature, Motion Histogram feature, and their combinations. The last column shows the accuracy on the cross-validation set with a 95\% confidence interval. 
If we do not use high-cloud segmentation, the combination of HOG and Motion Histogram features outperforms each of them. 
If we use high-cloud segmentation, the combination of these two features also performs better than each of them, and it also outperforms the combination of features without high-cloud segmentation. In conclusion, the effectiveness of our detection scheme is due to \textit{both} high-cloud segmentation process and weak classifiers built on shape and motion features.

\subsection{Detection Result}

\begin{figure}[!tb]
\begin{minipage}{0.6\linewidth}
\includegraphics[width=\textwidth,trim=0.3cm 0cm 0.3cm 0cm,clip]{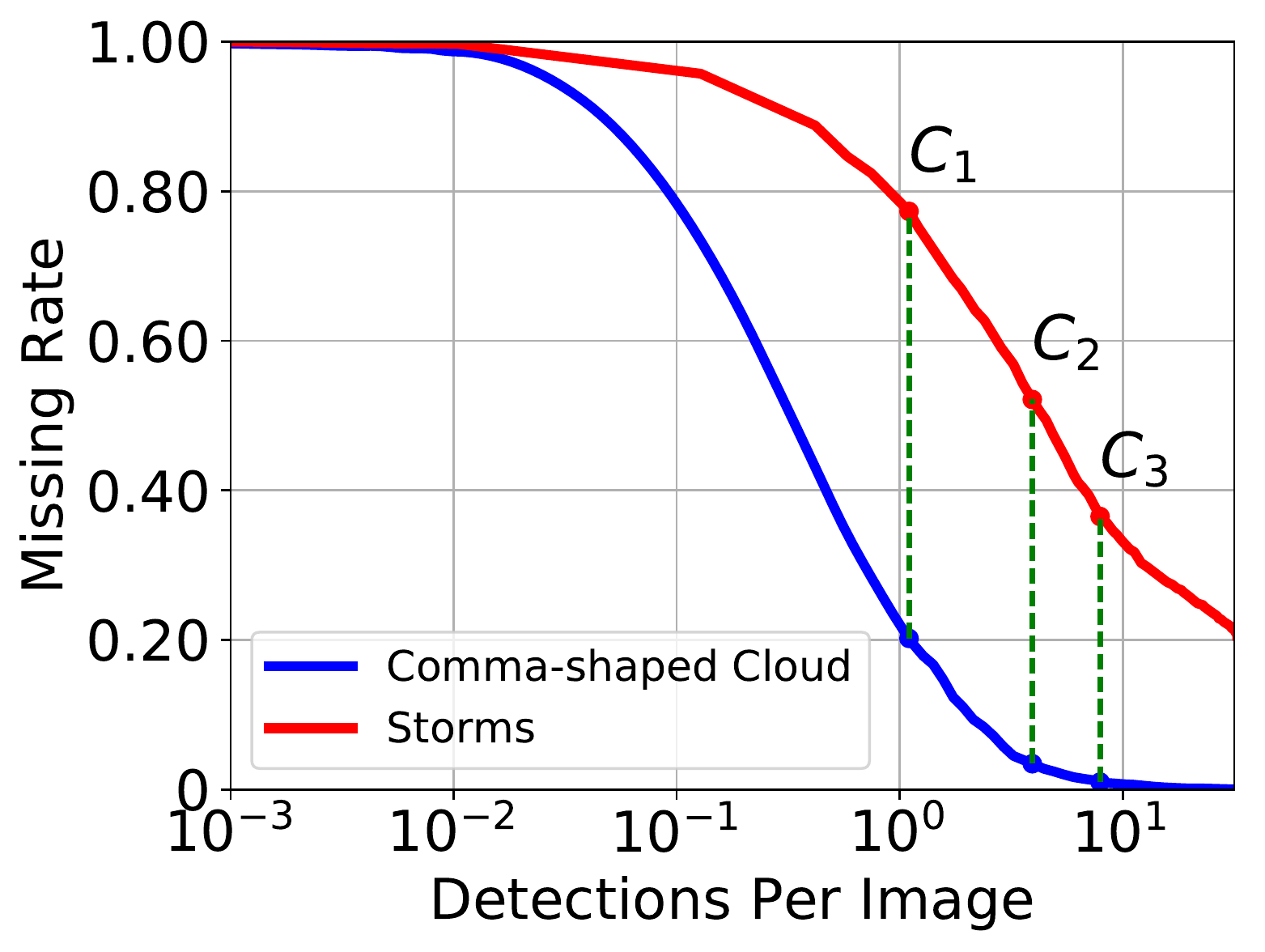}
\end{minipage}
\begin{minipage}{0.39\linewidth}
\begin{center}
\footnotesize
\begin{tabular}{p{4.1em}|p{0.7em}p{0.8em}p{0.8em}}
\multicolumn{4}{c}{REFERENCE POINTS}\\
\hline
Marker & $C_1$ &  $C_2$ & $C_3$ \\
\hline
Cutoff & 0.52 & 0.51 & 0.50\\
\hline
Detections & && \\
Per Image & 0.04 & 0.59 & 0.89\\
 (log 10)  & &&  \\
\hline
Missing & &&  \\
Rate (Clouds) & 0.20 & 0.03 & 0.01\\
\hline
Missing & &&  \\
Rate (Storms) & 0.77 & 0.52 & 0.36\\
\hline
\end{tabular}
\end{center}
\end{minipage}
\caption{Evaluation curves of our comma-shaped clouds detection method. \textit{Left}: Missing rate curve with Detections. \textit{Right}: Some reference cutoff values on the curve.}
\label{fig:missing-detect}
\end{figure}

The evaluation in Fig.~\ref{fig:missing-detect} shows our model can detect up to 99.39\% of the labeled comma-shaped clouds and up to 79.41\% of storms of the year 2011 and 2012. Here we define the term ``detect comma-shaped clouds'' as: If our method outputs a bounding box having IoU $\geq$ 0.50 with the labeled region, we consider such bounding box detects comma-shaped clouds; otherwise not. We also define ``detect a storm'' as: If any storm in the NOAA storm database is captured within one of our output bounding boxes, we consider we detect this storm. 

The comma-shaped clouds detector outputs the probability $p$ of each bounding box from AdaBoost classifier. If $p \geq p_0$, this bounding box consists of comma-shaped clouds. We recommend $p_0$ to be set in [0.50, 0.52], and we provide three reference probabilities $p_0$ = 0.50, 0.51 and 0.52. The number of detections per image as well as the missing rate of comma-shaped clouds and storms corresponding to each $p_0$ are available in the right part of Fig.~\ref{fig:missing-detect}. For a user who desires high recall rate, {\it e.g.} a meteorologist, we recommend setting $p_0$ = 0.50. The recall rate of the comma-shaped clouds is 99\% and the recall rate of storms is 64\% under that setting. Our detection method will output an average of 7.76 bounding boxes per satellite image. Other environmental data, like wind speed and pressure, are needed to be incorporated into the system to filter the bounding boxes. For a user who desires accurate detections, we recommend setting $p_0$ = 0.52. The recall rate of the comma-shaped clouds is 80\%, and our detector outputs an average of 1.09 bounding boxes per satellite image. The recall rate is reasonable, and the user will not get many incorrectly reported comma-shaped clouds.

The setting $p_0\in$ [0.50, 0.52] could give us the best performance for several reasons. When $p_0$ goes under the value 0.50, the missing rate of the comma-shaped clouds almost remains the same value ($\sim$1\%), and we need to check more than 8 bounding boxes per image to find these missing comma-shaped clouds. It consumes too much human effort. When $p_0$ goes over the value 0.52, the missing rate of comma-shaped clouds goes over 20\%, and the missing rate of storms goes over 77\%. Since missing a storm could cause severe loss, $p_0 > 0.52$ cannot provide us a recall rate that is high enough for the storm detection purpose.

Though our comma-shaped clouds detector can effectively cover most labeled comma-shaped clouds, it still misses at least 20\% storms in the record. Among different types of the storms, severe weather events on the ocean\footnote{Here severe weather events on the ocean includes marine thunderstorm wind, marine high wind, and marine strong wind.} have a higher probability to be detected in the algorithm than other types of severe weather events. At the point of the largest recall, our method detects approximately 85\% severe weather events on the ocean versus 75\% on the land. Our detector misses such events because severe weather does not always happen near the low center of the comma-shaped clouds. According to~\cite{carlson1980airflow} and~\cite{stewart1989winter}, the exact cold front and the warm front streamline cannot be accurately measured from satellite images. Hence, comma-shaped clouds are simply an indicator of storms, and further investigation in their geological relationships is necessary to improve our method.

\subsection{Storm Detection Ability}

We compare the storm identification ability of our algorithm with other baseline methods that use satellite images. 
The first baseline method comes from~\cite{morel2002climatology} and~\cite{fiolleau2013algorithm}, and the second baseline improves the first in~\cite{lakshmanan2003multiscale}. We call them~\textit{Intensity Threshold Detection} and~\textit{Spatial-Intensity Threshold Detection} hereafter.

The Intensity Threshold Detection uses a fixed reflectivity level of radar or infrared satellite data to identify a storm. A continuous region with a reflectivity level larger than a certain threshold $I_0$ is defined as a storm-attacked area. Spatial-Intensity Threshold Detection improves it by changing the cost function to be a weighted sum: 
\begin{equation*}
E = \sum_{i = 1}^{n} \lambda d_m\left(x_i\right) + \left(1 - \lambda\right) d_c\left(x_i\right), 
\end{equation*}
where $X = \left\{x_i\right\}_{i = 1}^{n}$ is the point set representing a cloud patch, $d_m$ is the spatial distance within the cluster, and $d_c$ is the difference between the pixel brightness $I\left(x_i\right)$ and the average brightness of the cloud $X$.

We make some necessary changes to the baselines to make two methods comparable. First, we explore different $I_0$ value, because we use the different channels and satellites with the baselines. In addition, the light distribution of images is changed through histogram equalization in the preprocessing stage, so we cannot simply adopt $I_0$ used in the baselines.
Second, we change the irregular detected regions to the square bounding boxes and use the same criteria to define positive and negative detections.
We adopt the idea in~\cite{lakshmanan2009efficient} and view these pixel distributions as 2D GMM. We use Gaussian means and the larger eigenvalue of Gaussian covariance matrix to approximate a bounding box center and a bounding box size, respectively. The number of Gaussian components and other GMM parameters are estimated by mean Silhouette Coefficient~\cite{rousseeuw1987silhouettes} and the k-means++ method.

\begin{figure}[tbp!]
\begin{minipage}{0.63\linewidth}
\centering
\includegraphics[width=\textwidth,trim=0cm 0.25cm 0cm 0cm,clip]{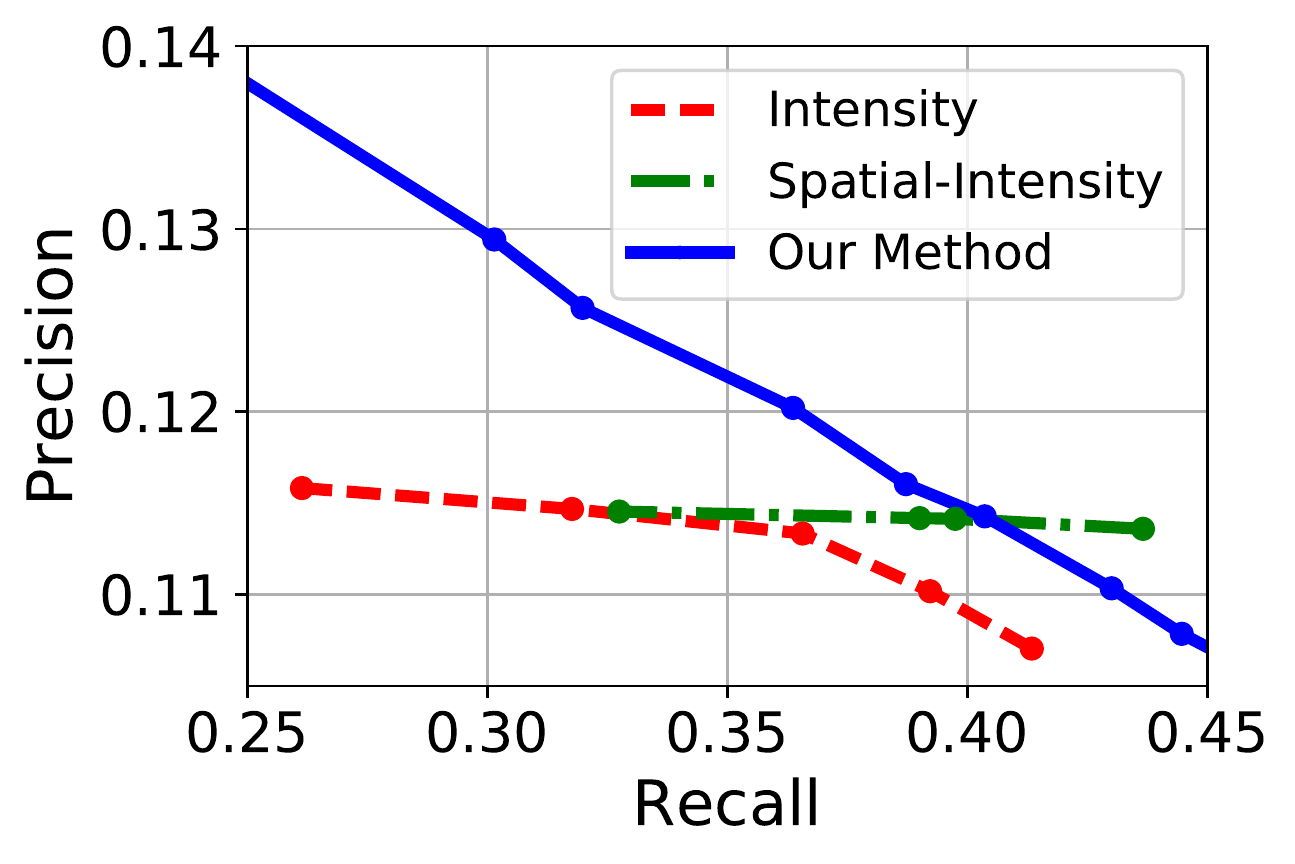}
\end{minipage}
\begin{minipage}{0.36\linewidth}
\begin{center}
\footnotesize
\begin{tabular}{c|c}
\multicolumn{2}{c}{MAXIMUM RECALL}\\
\multicolumn{2}{c}{OF STORMS}\\
  \hline
Method & Recall\\
 \hline
Intensity & 0.41  \\
 \hline
Spatial- & 0.44  \\
Intensity & \\
 \hline
\textbf{Our} & \textbf{0.79} \\
\textbf{Method} & \\
 \hline
\end{tabular}
\end{center}
\end{minipage}
\caption{Comparison of the baseline methods. \textit{Left}:
Part of Recall-Precision curve of the two baseline storm detection methods and our method. \textit{Right}: The maximum recall rate they can reach. Here Intensity = Intensity Threshold Detection and Spatial-Intensity = Spatial-Intensity Threshold Detection.}
\label{fig:baseline}
\end{figure}

The partial recall-precision curve in Fig.~\ref{fig:baseline} shows that our method outperforms both Intensity Threshold Detection and Spatial-Intensity Threshold Detection when the recall is less than 0.40. We provide only a partial recall-precision curve because of the limited range of $I_0$ under the limited time and computation resources. In our experiment, we change parameters $I_0$ in Intensity Threshold Detection from 210 to 230. When $I_0$ goes over 230, very few pixels would be selected so this method cannot ensure high recall rate. When $I_0$ goes under 210, many pixels representing low clouds are also included in the computation, which slows down the computations ($\sim$ 5 minutes per image). Consequently, we do not explore those values. As for Spatial-Intensity Threshold Detection, $I_0$ is fixed at the value of 225, and $\lambda$ is the weight between 0 and 1. When $\lambda$ changes from 0 to 1, the recall first goes up and then moves down, while the precision changes very little. The curve representing Spatial-Intensity Threshold Detection reaches the highest recall at 43.66\% when $\lambda$ approaches 0.7. 

Compared with the two baselines that detect storm events directly, our proposed method has the following strengths: (1) Our method can reach a maximum recall of 79.41\%, almost twice as those of the baseline methods. Due to computational speed issues, we could not increase the recall rate of the two baseline methods to be higher than 45\%, which limits their use in practical storm detections. For our method, however, we can reach a high recall rate without heavy computational cost.
(2) Our method outperforms these two baseline methods in the precision rate of Fig.~\ref{fig:baseline}. Compared with these two methods that mostly rely on pixel-wise intensity, our method comprehensively combines the shape and motion information of clouds in the system, leading to a better performance in storm detection.

None of the three curves in Fig.~\ref{fig:baseline} have a high precision of detecting storm events, because this task is difficult especially without the help of other environmental data.
In addition, our method aims to detect comma-shaped clouds, rather than to forecast storm locations directly. Sometimes severe storms are reported later than the appearance of comma-shaped clouds. Such cases are not counted in the precision rate of Fig.~\ref{fig:baseline}. In those cases, our method can provide useful and timely information or alerts to meteorologists who can make the determination.

Fig.~\ref{fig:baseline} also points out the importance of exploring the spatial relationship between comma-shaped clouds and storm observations, as the Spatial-Intensity Threshold Detection method slightly outperforms our method when the recall rate is higher than 0.40~. According to the trend of the green curve, adding spatial information to the detection system can improve the performance to some extent. We will consider combining spatial information into our detection framework in the future.

\begin{figure*}[!bthp]\centering
\includegraphics[width=.32\textwidth]{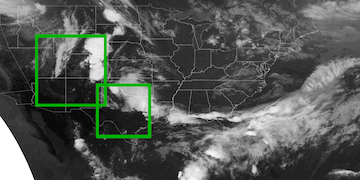}
\includegraphics[width=.32\textwidth]{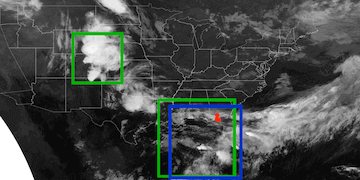}
\includegraphics[width=.32\textwidth]{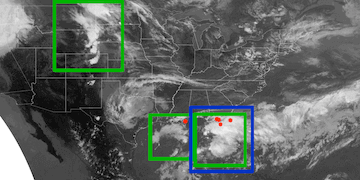}\\
\hfill ($a_1$) 00:45:19, June-07-2012 UTC \hfill ($a_2$) 05:15:18, June-07-2012 UTC \hfill ($a_3$) 19:45:19, June-07-2012 UTC \hfill\mbox{}\\
\includegraphics[width=.32\textwidth]{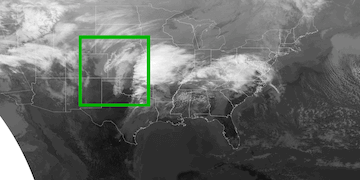}
\includegraphics[width=.32\textwidth]{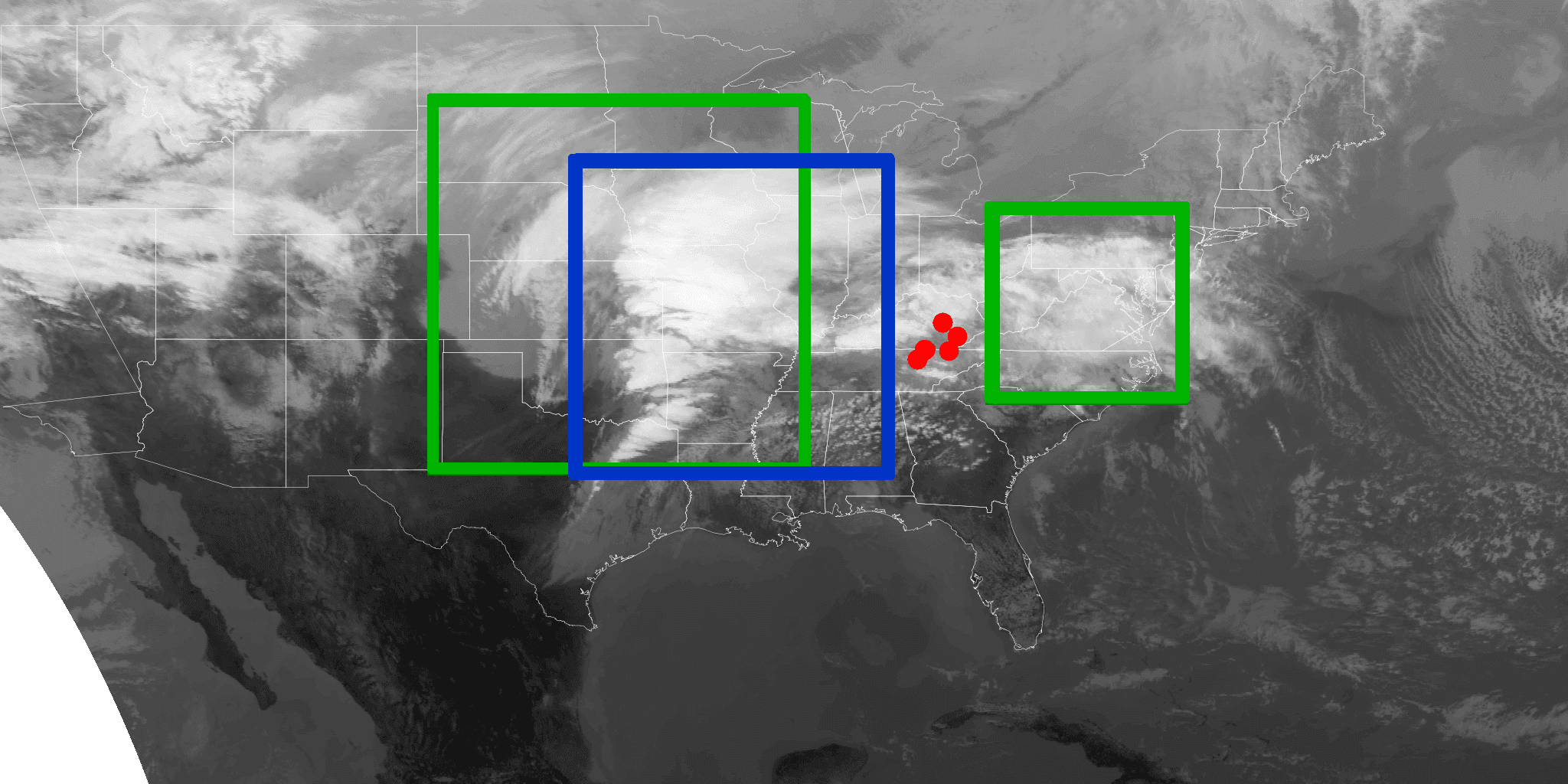}
\includegraphics[width=.32\textwidth]{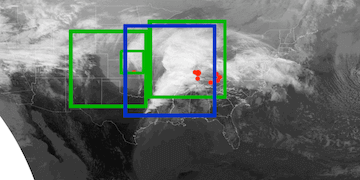}\\
\hfill ($b_1$) 16:15:17, Feb-24-2011 UTC \hfill ($b_2$) 18:15:19, Feb-24-2011 UTC \hfill ($b_3$) 21:45:19, Feb-24-2011 UTC \hfill\mbox{}\\
\includegraphics[width=.32\textwidth]{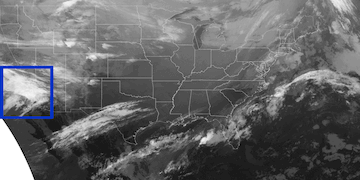}
\includegraphics[width=.32\textwidth]{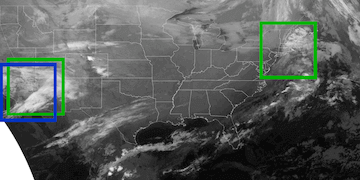}
\includegraphics[width=.32\textwidth]{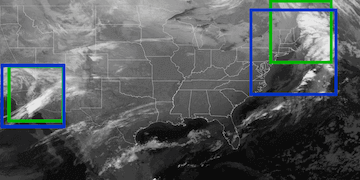}\\
\hfill ($c_1$) 21:15:19, Jan-02-2011 UTC \hfill ($c_2$) 02:15:19, Jan-03-2011 UTC \hfill ($c_3$) 05:45:18, Jan-03-2011 UTC \hfill\mbox{}\\
\caption{(a-c) Three detection cases. Green frames: our detection windows; Blue frames: our labeled windows; Red dots: storms. Some images have blank in left-bottom because it is out of the satellite range.}
\label{fig:casestudy}
\end{figure*}

\section{Case Studies}\label{sec:casestudy}

We present three case studies (a-c) in Fig.~\ref{fig:casestudy} to show the effectiveness and some imperfections of our proposed detection algorithm. The green bounding boxes are our detection outputs, the blue bounding boxes are comma-shaped clouds identified by meteorologists, and the red dots indicate severe weather events in the database~\cite{stormeventdatabase}. The detection threshold is set to 0.52 to ensure the precision of each output-bounding box. The descriptions of these storms are summarized from~\cite{annual2011summary}.

In the first case (row 1), strong wind, hail, and thunderstorm wind developed in the central and northeast part of Colorado, west of Nebraska and east of Wyoming, on late June 6, 2012. The green bounding box in the top-left corner of Fig.~\ref{fig:casestudy} - ($a_1$) indicated this region. Then, a dense cloud patch moved in the eastern direction and covered eastern Wyoming, western South Dakota, and western Nebraska on early June 7, 2012. At that time, these states reported property damages in different degrees. Later on June 7, 2012, the cloud patch became thinner as it moved northward to Montana and North Dakota, as shown in ($a_3$). Our method had a good tracking of that cloud patch all the time, even though the cloud shape did not look like a typical comma. In comparison, human eyes did not recognize it as a typical comma shape because it lost the head part. 
Another region detected to have a comma-shaped cloud in ($a_1$) was around North Texas and Oklahoma. At that time, hail and thunderstorm winds were reported, but the comma shape in the cloud began to disappear. Another comma-shaped cloud began to form in the Gulf of Mexico, as seen in the center part of ($a_1$). At that time, the comma shape was too vague to be discovered by either our computer detector or by human eyes. As time passed ($a_2$), the comma-shaped cloud appeared, and it was detected by both our detector and human eyes. The clouds gathered as some severe events in North Florida in ($a_2$). According to the record, a person was injured, and Florida experienced severe property damage at that time. Later that day, the large comma-shaped cloud split into two parts. The cloud patch in the west had an incomplete shape, which is difficult for human eyes to discover, as shown in ($a_3$). However, our method successfully detected this change. In addition, our method detected all the recorded severe weather events. This example indicates our method is able to detect incomplete or atypical comma-shaped clouds, including when one comma-shaped cloud splits into two parts.

In the second case (row 2), a comma-shaped cloud appeared in the sky over Oklahoma, Kansas, and Missouri on Feb 24, 2011, when these areas were attacked by winter weather, flooding, and thunderstorm winds. Our method detected the comma-shaped cloud half an hour earlier than human eyes were able to capture it, as shown in ($b_1$). Soon ($b_2$), a clear comma-shaped cloud formed in the middle of the image, which was detected by both our method and human eyes. Red dots in ($b_2$) show the location of some severe weather events happened in Tennessee and Kentucky at that time. Since the cloud patch was large, it was difficult to include the whole cloud patch in one bounding box. In that case, human eyes could correctly figure out the middle part of the wide cloud to label the comma-shaped cloud. In comparison, our detector used two bounding boxes to cover the cloud patch, as shown in ($b_2$) and ($b_3$). Because there was only one comma-shaped cloud, our method outputs a false negative in that case.

In the third case (row 3), there were two comma-shaped cloud patches from late Jan 2, 2011 to the early next day, located in the left and the right part of the image, respectively. Our method detected the comma-shaped cloud in south California one hour ({\it i.e.}, two continuous satellite images) later than the human eye detected it. Importantly, however, after the region is detected, our method detected the right comma-shaped cloud over the North Atlantic Ocean one hour earlier than human eyes did. As indicated in the left part of ($c_2$) and ($c_3$), our output is highly overlapped with the labeled regions. Our method was able to recognize the comma-shaped cloud when the cloud just began to form in ($c_2$). At the beginning, human eyes cannot recognize its shape, but our method was able to capture that vague shape and motion information to make a correct detection.

To summarize these studies, our method can capture most human-labeled comma-shaped clouds. Moreover, our method can detect some comma-shaped clouds even before they are fully formed, and our detections are sometimes earlier than human eye recognition. These properties indicate that using our method to complement human detections in practical weather forecasting may be beneficial. On the other hand, our detection scheme has a weakness as indicated in case (b). It has difficulty outputting the correct position of spatially wide comma-shaped clouds.

\section{Conclusions}~\label{sec:conclude}

We propose a new computational framework to extract the shape-aware cloud movements that relate to storms. Our algorithm automatically selects the areas of interest at suitable scales and then tracks the evolving process of these selected areas. Compared with human annotator's performance, the computational algorithm provides an objective (yet agnostic) standard for defining the comma shape. The system can assist meteorologists in their daily case-by-case forecasting tasks.

Shape and motion are two visual clues frequently used by meteorologists in interpreting comma-shaped clouds. Our framework includes both the shape and the motion features based on cloud segmentation map and correlation with motion-prior map. Our experiments also validate the usage of these two visual features in detecting comma-shaped clouds. Further, considering the high variability of cloud appearance in satellite images affected by seasonal, geographical and temporal factors, we take a learning-based approach to enhance the robustness, which may also benefit from additional data.

Finally, the detection algorithm provides us a top-down approach to explore how severe weather events happen. Our future work will integrate this framework with the use of other data sources and models to improve reliability and timeliness of storm forecasting.

\section*{Acknowledgment}

We thank the anonymous reviewers and the associate editor for their constructive comments. 
We thank the National Oceanic and Atmospheric Administration (NOAA) for providing the data. Yu Zhang, Yizhi Huang, and Jianyu Mao assisted with data collection, labeling, and visualization, respectively.
Haibo Zhang also provided feedback on the paper.

\ifCLASSOPTIONcaptionsoff
\newpage
\fi

\newcommand{\BIBdecl}{\setlength{\itemsep}{0.25 em}}
\bibliographystyle{IEEEtran}
\bibliography{weather}

\begin{IEEEbiography}[{\includegraphics[width=1in, height=1.25in,clip,keepaspectratio,trim={120 100 120 0}]{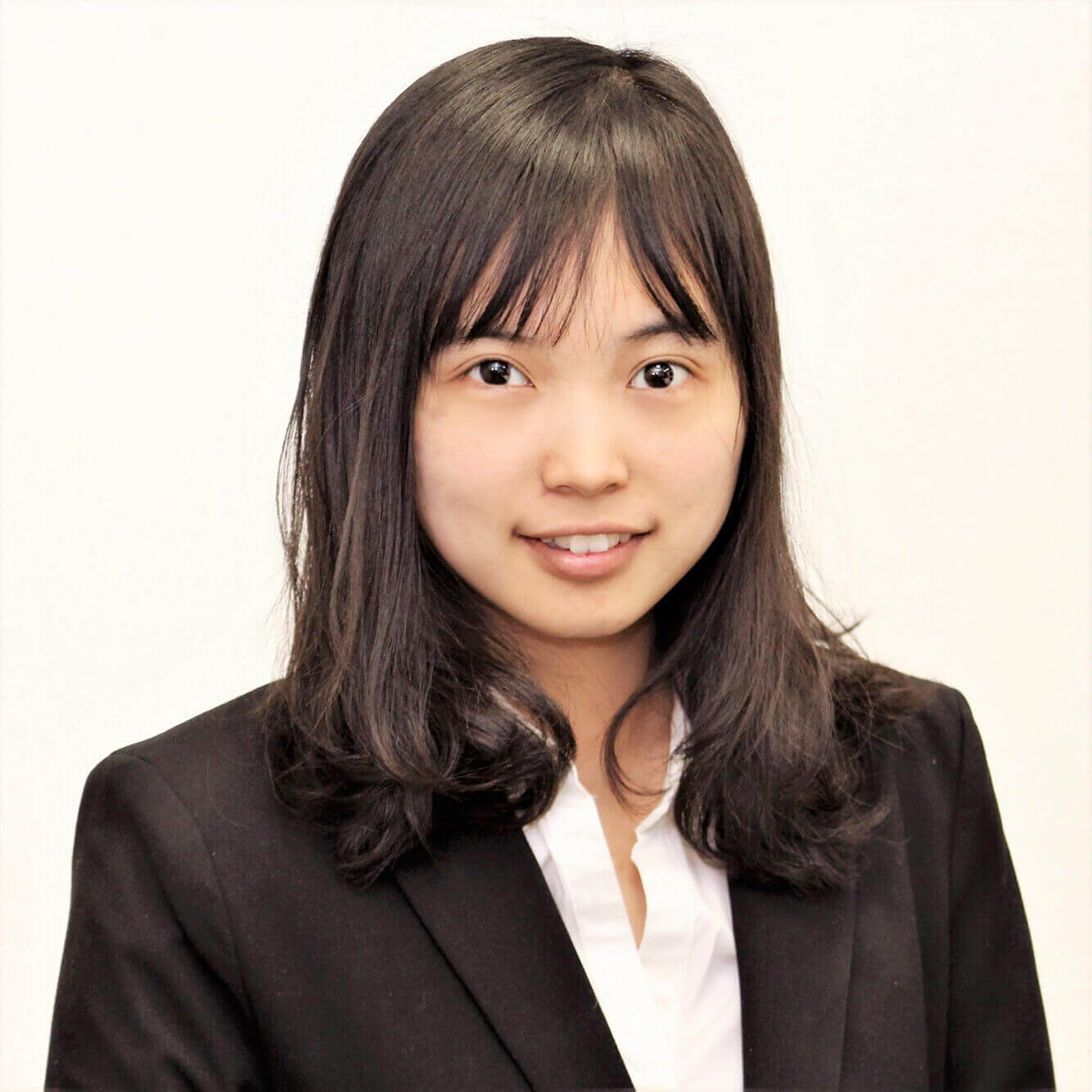}}]{Xinye~Zheng}
received the Bachelor's degree in Statistics from the University of Science and Technology of China in 2015. She is currently a PhD candidate and Research Assistant at the College of Information Sciences and Technology, The Pennsylvania State University. Her research interests include big visual data, statistical modeling, and their applications in meteorology, biology, and arts.
\end{IEEEbiography}

\vskip 0pt plus -1fil

\begin{IEEEbiography}[{\includegraphics[width=1in,height=1.25in,clip,keepaspectratio]{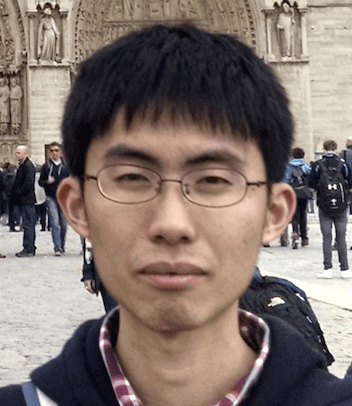}}]{Jianbo~Ye}
received the PhD degree in 
information sciences and technology
from The Pennsylvania State University in 2018. 
He received the Bachelor's degree in mathematics from the
University of Science and Technology of China in 2011. 
He served as a research postgraduate at
The University of Hong Kong (2011-2012), a research intern at Intel (2013), a research intern at Adobe (2017), and a research assistant at Penn State's College of Information Sciences and Technology and Department of Statistics (2013-2018).
His research interests include statistical modeling
and learning, numerical optimization and method, and affective image modeling. He is currently a scientist at the Amazon Lab126.
\end{IEEEbiography}

\vskip 0pt plus -1fil

\begin{IEEEbiography}[{\includegraphics[width=1in,height=1.25in,clip,keepaspectratio,trim={15 0 0 50}]{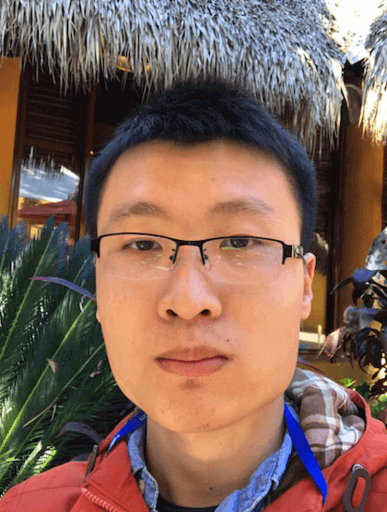}}]{Yukun~Chen}
received the Bachelor's degree in Applied Physics from the University of Science and Technology of China in 2014. He is currently a PhD candidate and Research Assistant at the College of Information Sciences and Technology, The Pennsylvania State University. He has been a summer intern at Google (2017) and Facebook (2018). His research interests include computer vision, multimedia, and machine learning.
\end{IEEEbiography}
\vskip 0pt plus -1fil

\begin{IEEEbiography} [{\includegraphics[width=1in,height=1.25in,clip,keepaspectratio,trim={0 45 0 0}]{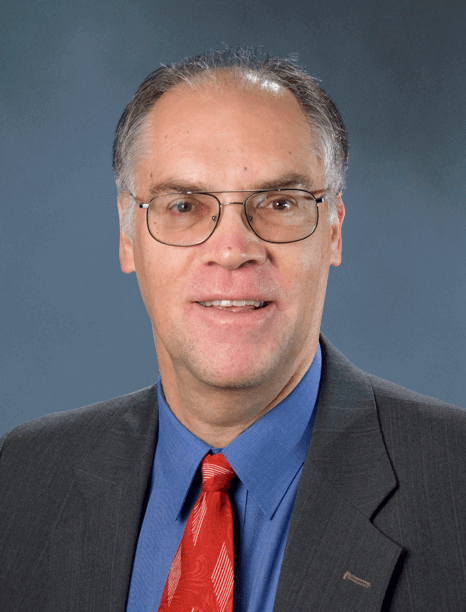}}] {Stephen~Wistar}
is a Certified Consulting Meteorologist (CCM) and Senior Forensic Meteorologist. He has worked on numerous cases involving Hurricane Katrina, building collapses, flooding and slip and falls. His work involves explaining meteorology to the non-scientist. At AccuWeather, Steve constructs impartial scientific weather evidence for use in courts of law to support the prosecution, which has aided Steve in testifying over 125 times in courtrooms or depositions. Furthermore, he coordinates numerous past weather research projects using
forensic meteorology, like the 250 reports Steve wrote on the impacts of Hurricane Katrina at specific locations on Gulf Coast.
\end{IEEEbiography}
\vskip 0pt plus -1fil

\begin{IEEEbiography}[{\includegraphics[width=1in,height=1.25in,clip,keepaspectratio]{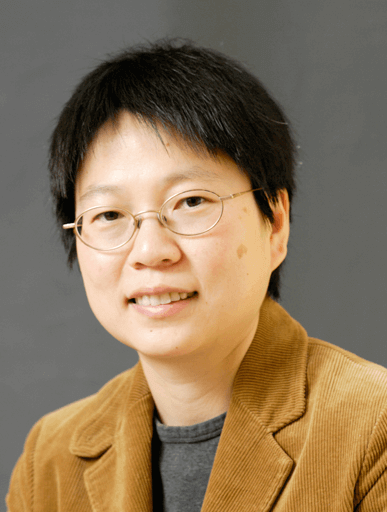}}]{Jia~Li}
is a Professor of Statistics at The Pennsylvania State University. She received the MS degree in Electrical Engineering, the MS degree in Statistics, and the PhD degree in Electrical Engineering, all from Stanford University. She worked as a Program Director in the Division of Mathematical Sciences at the National Science Foundation from 2011 to 2013, a Visiting Scientist at Google Labs in Pittsburgh from 2007 to 2008, a researcher at the Xerox Palo Alto Research Center from 1999 to 2000, and a Research Associate in the Computer Science Department at Stanford University in 1999. Her research interests include statistical modeling and learning, data mining, computational biology, image processing, and image annotation and retrieval.
\end{IEEEbiography}
\vskip 0pt plus -1fil

\begin{IEEEbiography}[{\includegraphics[width=1in,height=1.25in,clip,keepaspectratio]{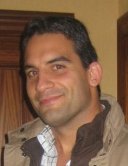}}]{Jose~A.~Piedra-Fern\'andez}
received the Bachelor’s degree in 
computer science and the M.S. and Ph.D. degrees from the University of Almería, Almería, Spain, in 2001, 2003, and 2005, respectively.
He is currently an Assistant Professor at the University of Almería. He visited the College of Information Sciences and Technology, The Pennsylvania State University, University Park, PA, USA, in 2008-2009. His main research interests include image processing, pattern recognition, and
image retrieval. He has designed a hybrid system applied to remote sensing problems.
\end{IEEEbiography}
\vskip 0pt plus -1fil

\begin{IEEEbiography} [{\includegraphics[width=1in,height=1.25in,clip,keepaspectratio]{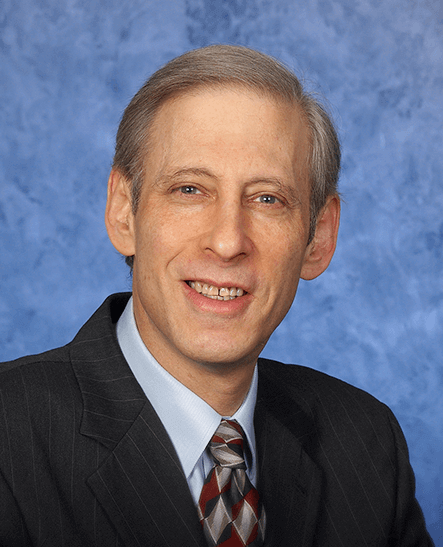}}] {Michael~A.~Steinberg}
received the BS in atmospheric sciences from Cornell University and the MS degree in meteorology from The Pennsylvania State University. He is an Expert Senior Meteorologist, Senior Vice President and Emeritus member of the Board of Directors of AccuWeather, Inc., where he has been employed since 1978. In this role, he interacts in a wide variety of scientific, tactical and strategic areas. He is a Fellow of the American Meteorological Society (AMS) and was a recipient of their 2016 Award for Outstanding Contribution
to the Advance of Applied Meteorology for numerous, visionary innovations and accomplishments in meeting public and industrial needs for weather information. He has been a recipient of research grants from the NASA Small Business Innovation Research program and the Ben Franklin/Advanced Technology Center of Pennsylvania, and is the inventor or co-inventor on numerous patents related to weather indices and location-based services.\end{IEEEbiography}
\vskip 0pt plus -1fil

\begin{IEEEbiography}[{\includegraphics[height=1.3in,clip,keepaspectratio,trim={2 7 2 0}] {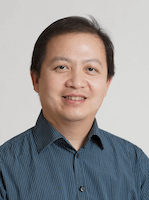}}]{James~Z.~Wang}
is a Professor of Information Sciences and Technology at The Pennsylvania State University. He received the bachelor’s degree in mathematics and computer science {\it summa cum laude} from the University of Minnesota, and the MS degree in mathematics, the MS degree in computer science and the PhD degree in medical information sciences, all from Stanford University. His research interests include image analysis, image modeling, image retrieval, and their applications. He was a visiting professor at the Robotics Institute at Carnegie Mellon University (2007-2008), a lead special section guest editor of the IEEE Transactions on Pattern Analysis and Machine Intelligence (2008), and a program manager at the Office of the Director of the National Science Foundation (2011-2012). He was a recipient of a National Science Foundation Career award (2004).
\end{IEEEbiography}
\end{document}